\documentclass[]{interact}

\usepackage{epstopdf}
\usepackage[caption=false]{subfig}

\usepackage[numbers,sort&compress]{natbib}
\bibpunct[, ]{[}{]}{,}{n}{,}{,}
\renewcommand\bibfont{\fontsize{10}{12}\selectfont}
\makeatletter
\def\NAT@def@citea{\def\@citea{\NAT@separator}}
\makeatother

\theoremstyle{plain}

\theoremstyle{definition}

\theoremstyle{remark}

\usepackage{amsmath,amsfonts,amssymb}
\usepackage{algorithmic}
\usepackage{array}
\usepackage{textcomp}
\usepackage{verbatim}
\usepackage{graphicx}

\usepackage{xcolor}
\usepackage[normalem]{ulem}

\newif\ifreview
\reviewfalse 

\ifreview
  \newcommand{\revadd}[1]{\textcolor{blue}{#1}}
  \newcommand{\revdel}[1]{\textcolor{red}{\sout{#1}}}
\else
  \newcommand{\revadd}[1]{#1}
  \newcommand{\revdel}[1]{}
\fi

\usepackage[whole]{bxcjkjatype}
\usepackage{breqn}
\usepackage[dvipsnames,table,svgnames]{xcolor}
\usepackage{booktabs}\newcommand {\figref}[1] {Fig.~\ref{#1}}

\begin{document}

\articletype{FULL PAPER}

\title{Learning to Predict Contact Force Distributions from Vision Leveraging Object Geometry Priors}

\author{
\name{Ryo Hanai\textsuperscript{a}\thanks{CONTACT R. Hanai. Email: ryo.hanai@aist.go.jp}, Yukiyasu Domae\textsuperscript{a}, Ixchel G. Ramirez-Alpizar\textsuperscript{b}, Abdullah Mustafa\textsuperscript{a}, Floris Erich\textsuperscript{a} and Tetsuya Ogata\textsuperscript{a,c}
}
\affil{\textsuperscript{a}Artificial Intelligence Research Center, National Institute of Advanced Industrial Science and Technology (AIST), Tokyo, Japan;
\textsuperscript{b}Intelligent Systems Research Institute, National Institute of Advanced Industrial Science and Technology (AIST), Tokyo, Japan;
\textsuperscript{c}Institute of AI and Robotics, Graduate School of Fundamental Science and Engineering, Waseda University, Tokyo, Japan}
}

\maketitle

\begin{abstract}
Based on vision and prior experience, humans can make rough physical predictions and adjust their manipulation strategies. This paper aims to endow robots with a similar ability. To collect paired data of vision and forces, we use a rigid-body simulator commonly adopted in robotics. However, unlike simulators that output noisy point forces, humans are able to make \revdel{stable}\revadd{consistent} predictions even in unfamiliar situations. 
Based on this observation, we hypothesize that predicting smooth force distributions rather than raw point forces can improve both force prediction itself and downstream task performance.
To validate this hypothesis, we construct a model that predicts three-dimensional force distributions from a single RGB image of piled daily objects. The target distribution is generated by applying statistical smoothing to point forces obtained from the simulator. Moreover, by incorporating object geometry into the smoothing process, we aim to account for variations in contact states and achieve more \revdel{stable}\revadd{consistent} vision-based predictions.
We conduct extensive evaluations in both simulation and real environments. Results show that our approach improves prediction accuracy, enhances downstream task performance through smoothing, and further benefits from geometry-guided smoothing. Remarkably, the trained model generalizes effectively to real-world scenes despite being trained solely in simulation.
\end{abstract}


\section{Introduction}

Humans can make rough physical predictions of unknown objects and novel scenes based on visual information and accumulated experience, and can flexibly adapt their manipulation strategies using these predictions. For instance, when attempting to retrieve an object located at the bottom of a pile, a person naturally anticipates the downward forces from the objects above and chooses to extract the target horizontally.

Vision-based force prediction offers the advantage of providing non-contact estimates of forces across a wide range of objects and scenes. However, the problem is inherently ill-posed: high accuracy cannot always be achieved, and predictions can easily be misled. For example, objects with identical appearances but different physical properties may generate incorrect predictions, as can containers whose contents are visually hidden. Nevertheless, even if accurate prediction is not always feasible, approximate estimates are still valuable. They can be used to restrict the possible states of the environment, or, though not addressed in this study, to improve accuracy by combining vision with additional modalities such as language or tactile/force sensing. Based on this observation, this paper aims not to achieve exact predictions for specific objects, but rather to provide robots with the ability to make coarse yet useful predictions across diverse piled scenes, including those with unknown objects.

To this end, we propose a framework that enables robots to learn rough force prediction from vision. Specifically, we construct an end-to-end model trained on data collected through physics simulation. Recent progress in computer vision has shown that large datasets can support advanced prediction from visual input. However, to the best of our knowledge, no dataset currently exists that captures the forces arising from multi-object interactions in cluttered environments. Simulation thus becomes a practical alternative, as it provides direct access to information that is nearly impossible to measure in reality, such as the inter-object forces within piles of objects.

Although advanced simulation techniques, such as deformable-body simulation, can be employed, performing \revdel{stable}\revadd{consistent} simulations for a wide variety of objects and handling different types of deformations is challenging. Adjusting the models for each object and simulator requires considerable effort. As an alternative, we employ rigid-body simulation, which has been widely used in robotics and provides stable results in multi-body contact situations. This study aims to predict useful information for manipulation from the forces obtained through such simulators. However, one problem arises: these simulators output point forces at contact points, whereas in reality, contact is not always point-like. Objects can contact along surfaces and lines. Furthermore, point forces are highly sensitive to fine contact states and the mesh structure of the objects, making direct prediction prone to overfitting and difficult to generalize to novel objects and scenes.

A previous work~\cite{IROS2023:hanai} suggested that statistical smoothing could be effective to fill the gap between the point contacts in the simulator and non-point contacts in the real world. 
However, it could not account for differences in contact patterns, such as surface contact. 
\revadd{It relies on isotropic smoothing of point-contact forces. When attempting to redistribute sparse point forces to cover surface or line contacts, it tends to spread forces not only along the contact area but also in the normal direction. This can lead to undesired expansion of the force distribution into neighboring regions, including contacts
belonging to other objects.} Moreover, \revdel{it}\revadd{the previous work} did not analyze the effect of smoothing or evaluate the performance of downstream tasks in a ``real environment''.
In this work, we propose a new smoothing approach that incorporates a prior on the object geometry of the contacting object. By doing so, we expect the predicted force pattern to become more consistent with human intuition and the variation of physical contacts, such as point contact or surface contact, resulting in higher performance in downstream tasks.
\revadd{Although the supervision signal is designed to mitigate the gap between point-contact representations in simulation and surface/line contacts in the real world through smoothing, the model is trained to reproduce a simulator-consistent force distribution derived from the simulator, rather than to estimate physically accurate contact forces in the real world.}
To understand how \revdel{rough}\revadd{such} predictions can be effective, we conduct an extensive evaluation from two perspectives: (1) the relationship between smoothing methods and smoothing strength, and the generalization ability of the trained model, and (2) the performance of downstream tasks.

The main contributions of this study are as follows.
\begin{itemize}
 \item We propose a method to predict a rough force distribution from a single RGB image by taking into account the geometry of the contacting objects.
 \item We investigate how the smoothing method and its strength affect the prediction accuracy of force prediction on unknown scenes composed of unseen objects.
 \item We propose a method to evaluate the usefulness of the predicted force distributions using a downstream task. Specifically, we consider a task in which a target object is lifted from a pile and propose evaluation methods that utilize visual tracking in real environments, as well as metrics to assess the disturbances caused by surrounding objects.
 \item We also demonstrate that the trained model generalizes to real-world environments by evaluating its performance on downstream tasks.
\end{itemize}

\section{Related Work}
Studies on force prediction using vision have been conducted in various contexts.
The most common is the interaction between an object and a human: the contact force applied by the human hand to an object during object manipulation~\cite{Ehsani2020UseTF,8085141},
and the interaction forces between the human body and objects or the environment~\cite{Zhu2016InferringFA,Li2019Estimating3M}.
 Shin et al. estimated the interaction force applied to objects assuming a situation 
where the robot gripper applied a force~\cite{Shin2018SequentialIA}.
Im2Contact~\cite{Im2Contact} handles the prediction of extrinsic contact caused by a grasped object and another object from vision.
However, to the best of our knowledge, there have been no studies on the estimation of the contact force from the vision for a static scene with multiple objects interacting with each other, including objects that the robot is not touching.

The approaches for learning models to predict force from vision are divided into two categories: simulation-based approaches~\cite{Ehsani2020UseTF,Zhu2016InferringFA,Im2Contact,Wang2022VisualHR} and real data-based approaches~\cite{8085141,Shin2018SequentialIA,Collins2023VisualCP}.
Collecting data in a real-world environment requires considerable effort.
Collins et al. leveraged weakly labeled data in uncontrolled settings~\cite{Shin2018SequentialIA}.
Our method is simulation-based. Many simulation-based approaches aim to achieve accurate modeling.
However, the scenes targeted in this study have various appearances, geometries, and physical properties. Large datasets are required to achieve highly accurate predictions in such scenarios. In addition, there is inherent uncertainty.
Therefore, we aim to learn a model that can make predictions on real data, which have a different distribution from that of the simulation data, and generalize it to unseen objects by predicting rough patterns using data that are not large in scale.
To make these predictions, we are inspired by Lee et al.~\cite{Lee2020LearningQL}, although the problems are different. 
They trained the controller of a quadrupedal robot using only rigid terrain and a small set of procedurally generated terrains. However, the controller handled real complex terrains, including deformable terrain. 
We considered whether a rigid-body simulator that does not necessarily output a pattern close to the real force could be used to make useful predictions in the real world.

Visual information correlated with the force is necessary to estimate the interaction force from vision.
Many of the aforementioned studies used this movement~\cite{Ehsani2020UseTF,8085141,Li2019Estimating3M}
and/or deformation~\cite{Zhu2016InferringFA,Shin2018SequentialIA,Wang2022VisualHR}
as such visual cues.
However, these cues are unavailable for static scenes with no observable deformations.
Even in such cases, the prediction of physical properties is possible by assuming a statistical correlation between the object type, appearance, and physical properties such as mass.
Such an assumption may not be sufficient for a highly accurate estimation; however, it can be utilized for approximate inference.
For example, estimating tactile properties~\cite{Takahashi2018DeepVL},
stiffness~\cite{doi:10.1080/01691864.2022.2078669} by using vision has been explored.
Our approach involves end-to-end learning from appearance to physical interactions.
Although this approach does not explicitly estimate physical properties,
it is thought to internally correlate with the appearance of objects.

This study addresses the problem of estimating the interaction forces between a group of objects stacked according to gravity. This is a scene-understanding problem.
The human cognitive ability to perform visual physical reasoning of objects is called intrinsic physics, which has been studied in conjunction with the advancement of machine learning~\cite{Duan2022ASO}.
Stability prediction is a typical task in the physical reasoning of static scenes~\cite{Lerer2016LearningPI,Groth2018ShapeStacksLV}.
These studies included determining the stability of a scene with objects of various shapes stacked on top of each other, predicting the point of stability as a heat map, and predicting the outcomes of physical interactions. Motoda et al. applied stability prediction to bimanual picking~\cite{motoda:collapse}.
These studies differed from ours in that they did not directly predict the interaction force.

\section{Method}

\subsection{Overall Framework}

\begin{figure}[htbp]
\centerline{\includegraphics[width=\textwidth]{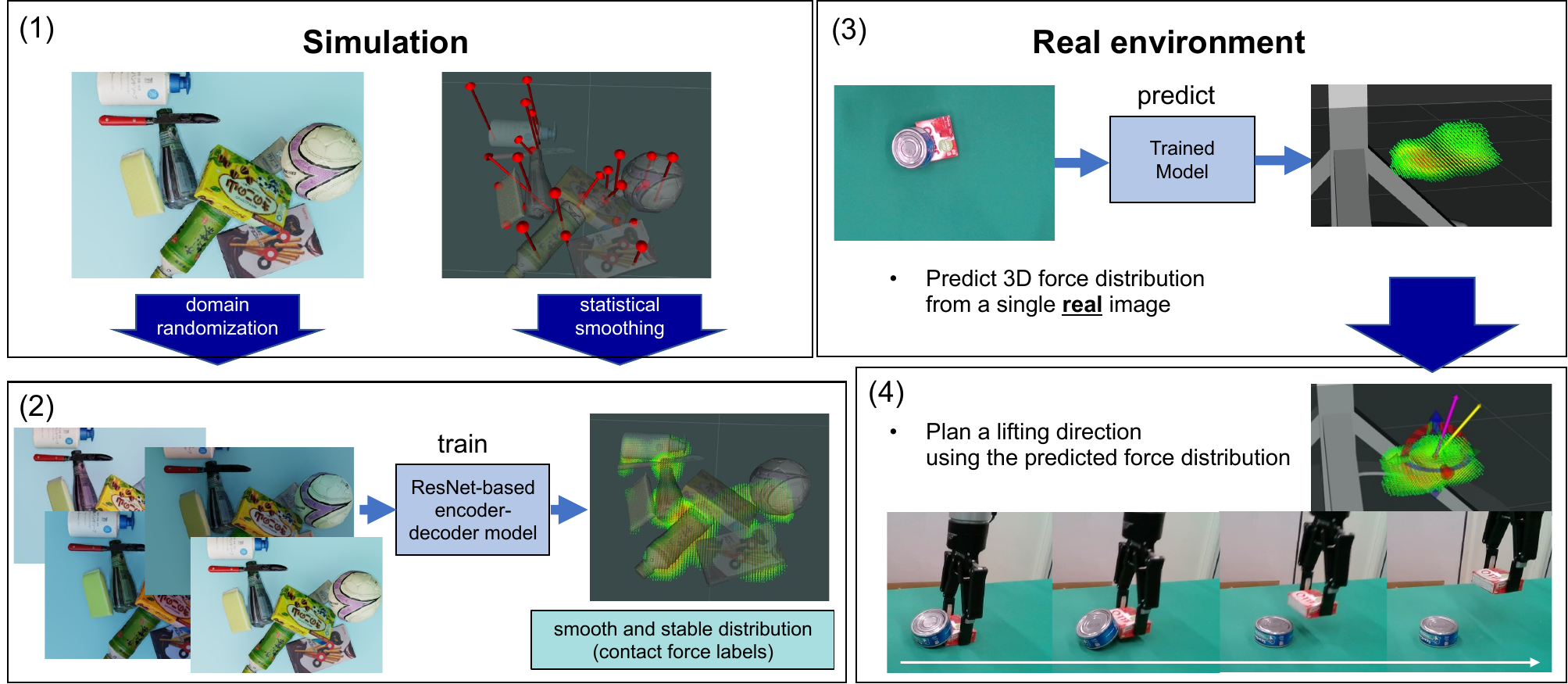}}
\caption{Overview of the proposed approach.
We train a model that predicts an approximate distribution of contact forces from a single image. (1) Paired data of an image and corresponding contact forces are generated by simulation. The contact force data are converted to a smooth and \revdel{stable}\revadd{consistent} distribution. 
(2) Training is performed using only the synthetic data.
(3) Use of the trained model to predict an approximate force distribution for real scenes.
The trained model generalizes to real-scene images because domain randomization is applied to generate the training images.
(4) Using the predicted force distribution, a lifting direction for picking a specified object is computed so as not to disturb the arrangement of surrounding objects.}
\label{fig:overview}
\end{figure}

We train a model end-to-end to predict a 3D force distribution from a single image on the dataset generated in simulation. \revadd{In this work, the model predicts only the force magnitude, rather than a full force vector field.
The force direction used for lifting planning in Section \ref{subsec:lifting_planning} is subsequently approximated from this magnitude.}
\figref{fig:overview} shows the overall framework.
First, the paired data of the images and corresponding contact forces are generated using a simulator.
The domain gap between synthetic and real images is reduced via domain randomization~\cite{Lee2021BeyondPT,Tobin2017DomainRF}.
Using these data, we train a model to predict the rough force distribution from a single RGB image.
The trained model is then used to predict force in a real environment.
Then, the predicted force distribution is used for downstream tasks, such as planning the lifting direction.
The force prediction model is identical to that used in our previous work~\cite{IROS2023:hanai}.
The force distribution is represented as a 3D voxel map. 
We interpret the prediction of the force distribution as an image-to-image translation problem by slicing the output force distribution in the z-axis iso-surface.
A ResNet-based encoder-decoder model is used for the translation. 
The encoder employs the feature extraction part of ResNet50~\cite{He2015DeepRL}.
The decoder comprises the residual blocks used by the Residual U-Net~\cite{Zhang2017RoadEB}.
The encoder is initialized with the pre-trained weight of ImageNet~\cite{Russakovsky2014ImageNetLS} and fine-tuning is performed.
The prediction model is trained using only synthetic data (single RGB and contact forces) obtained from a rigid body simulation. 

Rigid-body simulation is a well-established and stable technology in multibody interaction situations. 
Additionally, the effort required to create models for the simulation is relatively small, even when a large variety of objects is involved.
However, there is a difference between the simulated and actual forces.
The simulated forces are sparse point forces and susceptible to changes in the contact state and mesh structure.
Whereas the real force is not a clean contact caused by deformation, and contact sometimes occurs on a surface.
To alleviate this difference, we statistically convert the simulated point forces to a smooth distribution and use them as prediction targets.

\subsection{Geometry-aware Statistical Smoothing of Force Labels}

\begin{figure}[htbp]
 \centerline{\includegraphics[width=\columnwidth]{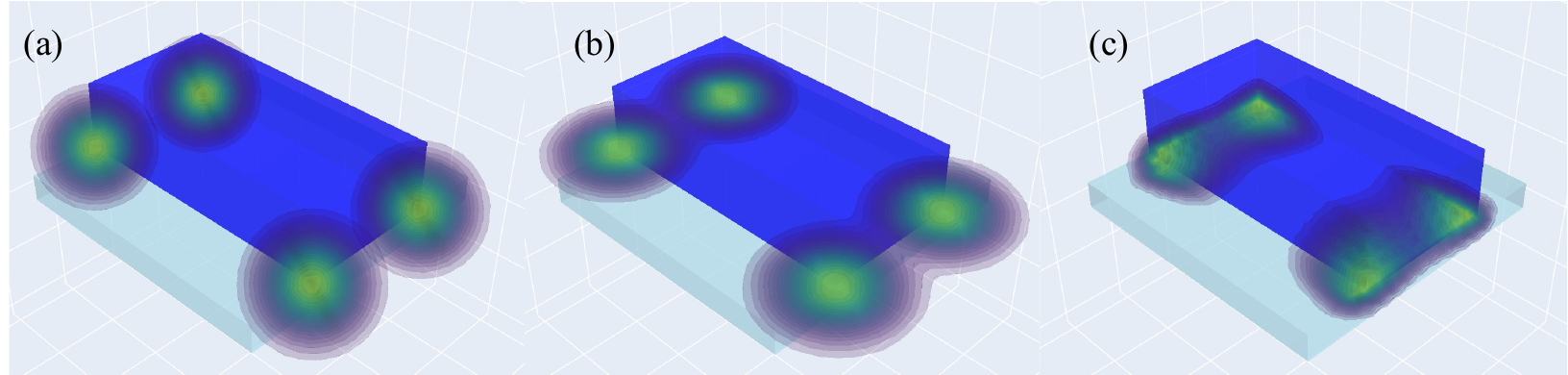}}
\caption{Different smoothing approaches: (a) isotropic, (b) covariance-based, and (c) geometry-aware smoothing.}
\label{fig:different_smoothing_approaches}
\end{figure}

\begin{figure}[htbp]
 \centerline{\includegraphics[width=0.7\columnwidth]{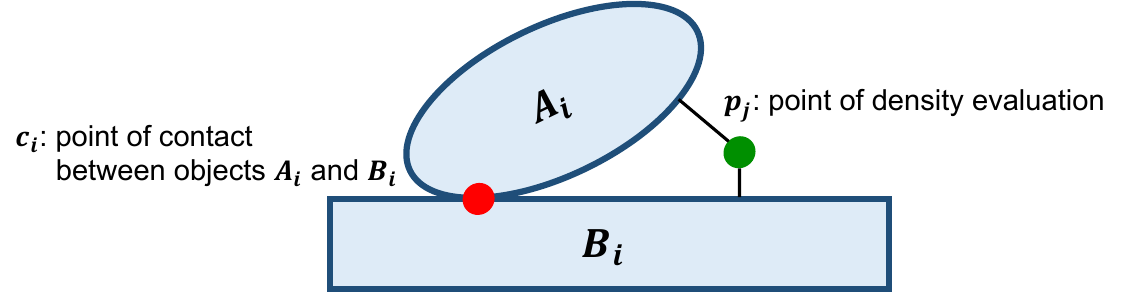}}
\caption{Geometry-aware smoothing.}
\label{fig:geometry_aware_smoothing}
\end{figure}

In a previous study, we isotropically smoothed force labels by applying weighted Kernel Density Estimation (KDE)~\cite{Chen2017ATO} in three-dimensional (3D) space.
Let the magnitude of the force at $n$ contact points $\bf{x_i}$ be $f_i$.
The voxel spacing is $h$. The force density at each voxel $\bf{x}$ is expressed as:
\begin{dgroup}
  \begin{dmath}
    \widetilde{f}({\bf x}) = \frac{1}{nh^3}\sum^{n}_{i=1}f_iK(\frac{{\bf x}-{\bf x_i}}{\sigma})
  \end{dmath}
  \begin{dmath}
    K({\bf x}) = \frac{1}{(2\pi)^{3/2}}e^{-\frac{{\bf \|x\|}^2}{2}},
  \end{dmath}
\end{dgroup}
where $K({\bf x})$ is the Gaussian kernel and ${\sigma}$ is the bandwidth parameter that controls the amount of smoothing. Because our objective is not to estimate the probability density, $\widetilde{f}({\bf x})$ is not normalized.
However, this method yields large force prediction errors, and the lifting directions computed from it are \revdel{unstable}\revadd{inconsistent} when the real-world contact is far from a pure point contact. 
To alleviate this problem, we aim to facilitate the prediction of force patterns according to contact states, such as point contact and surface contact. 
\figref{fig:different_smoothing_approaches} shows the idea.
In the figure, the point contact forces are applied at the corners of the rectangular object.
(a) When isotropic smoothing is applied, the generated distribution spreads significantly
in the direction perpendicular to the contact surface.
One approach to consider the spread of the contact force data generated by the simulator involves computing the covariance of the point force output by the simulator.
However, computing their covariances in a computationally stable manner is difficult because the simulated forces are extremely sparse.
(b) Instead of estimating the data distribution of the point forces, we use the object geometry as a potential prior to the forces.
However, this remains a challenge. 
Even if the covariance of the local geometry around a contact point is used, the estimated distribution spreads outside the contact area, as shown in (b).
This effect cannot be ignored because the contact points are often located near the boundary of the contact surface. 
(c) To avoid this problem, we propose a method for smoothing the point forces by multiplying the weights that attenuate according to the distance from the contacting objects.
\figref{fig:geometry_aware_smoothing} shows the concept, and the following formula is used to calculate the force density in each voxel.
\begin{dgroup}\label{eq:gafs}
    \begin{dmath}
        \widetilde{f}(p_j) = \sum_i\alpha_{i}f_{i}\widetilde{K}(p_j, c_i)
    \end{dmath}
    \begin{dmath}
        \widetilde{K}(p_j, c_i) = K(\frac{D(p_j, Ai)}{\sigma_g})K(\frac{D(p_j, Bi)}{\sigma_g})K(\frac{\|p_j-c_i\|_2)}{\sigma_f}) \label{eq:geo_weighting}
    \end{dmath}
    \begin{dmath}
        \alpha_i = 1/\sum_j\widetilde{K}(p_j, c_i)
    \end{dmath}
\end{dgroup}
where $f_i$ is the magnitude of the contact force at contact point $c_i$ (output of the simulator), $K$ is the Gaussian function, $A_i$ and $B_i$ are the regions occupied by the two objects in contact at contact point $c_i$, $D(p_j, A_i)$ is the distance between point $p_j$ and area $A_i$, and $\|\cdot\|_2$ denotes L2 norm, $\sigma_g$ is a parameter that adjusts the degree of attenuation with the distance from the object, and $\sigma_f$ is a parameter that adjusts the degree of force smoothing. 
In Equation \ref{eq:geo_weighting}, the first and second terms on the right-hand side represent the weights of the attenuation depending on the distance from the contacting object.
$\alpha_i$ is a normalization factor that preserves the force generated by contact before and after smoothing. 
Usually, a simplified collision mesh is used in physical calculations; however, the use of a visual mesh for smoothing facilitates the consideration of a more detailed geometry.
This smoothing method can be considered a natural extension of force label smoothing in previous work, which used no geometry-aware weights.
\revadd{In the case of surface contact, the distance to both contacting objects becomes zero on the contact surface. As a result, smoothing is fully applied along the contact surface, allowing the distribution to spread over a wide area. In contrast, along the contact surface normal direction, the distance to at least one of the objects increases, causing the smoothing effect to decay rapidly. Consequently, the distribution spreads primarily along the surface.
On the other hand, in point contact, any point in the vicinity of the contact location is separated from at least one of the objects. Therefore, the smoothing effect decays even near the contact point, resulting in a distribution that remains localized around the point.}
\revdel{This allows smoothing according to the contact state and mitigates the problem of distribution leakage outside the contacting objects.}
To efficiently evaluate the above equations, a voxel-based signed distance function (Voxel-based SDF)~\cite{zeng20163dmatch} is used.
SDF represents a shape’s surface by using a continuous volumetric field. 
The magnitude of a point in the field represents the distance to the surface boundary, and the sign indicates whether the region is inside (-) or outside (+) of the shape. We used DeepSDF~\cite{Park2019DeepSDFLC} to compute the SDF from a mesh.

\subsection{Force Label Scaling and Normalization}

There are various objects in our daily environment, ranging from light objects, such as sponges, to heavy objects, such as plastic bottles filled with content or metal blocks. Thus, the force has a wide range of values.
When predicting the forces in these situations, the smaller the applied force, the more important the small differences are.
This is also true for planning the lifting directions, which will be discussed later. 
It is expected to correctly predict patterns, such as the relative magnitudes of the forces.
Based on this consideration, the force distribution is converted to a log scale and normalized to $[0,1]$ within the upper and lower bounds of the training data. The loss is evaluated using Mean Squared Error (MSE).
This can be regarded as equivalent to structure-aware loss~\cite{Collins2023VisualCP} in that it incurs a large penalty for small errors.

\section{Experiments}

We focus on three aspects and design experiments to answer the following research questions.
\begin{description}
 \item[\bf RQ1:] Does the smoothed distribution make consistent predictions easier compared to the point forces?
 \item[\bf RQ2:] Is the predicted distribution useful for manipulation tasks?
 \item[\bf RQ3:] Does the prediction model transfer well to real-world scenes?
\end{description}

All experiments are conducted using piled scenes. To answer {\bf RQ1}, we train models that predict targets with different smoothing methods applied, and compare their prediction errors on unseen scenes.
To answer {\bf RQ2}, we consider a downstream task called lifting direction planning. We perform lifting on unseen scenes generated in simulation and compare the amount of disturbances caused to the surrounding objects while lifting. To answer {\bf RQ3}, we perform lifting from piled scenes in a real environment, and evaluate the resulting disturbances using vision-based 6DoF pose estimation.

\subsection{Dataset generation in simulation}

\begin{figure}[tbp]
 \centerline{\includegraphics[width=0.8\columnwidth]{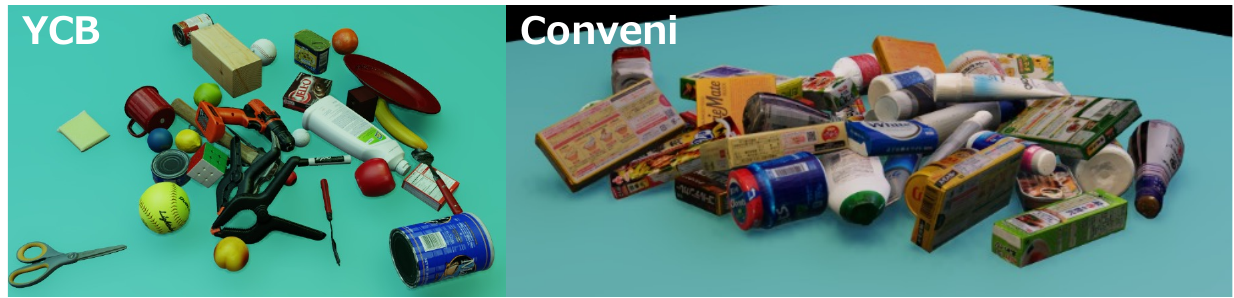}}
\caption{Convenience store items included in the dataset.}
\label{fig:conveni_items}
\end{figure}
\begin{table}[tbp]
\caption{YCB+Conveni Dataset}
\label{tabl:num_3dmodels}
\setlength{\tabcolsep}{3pt}
\begin{center}
{\scalebox{0.8}{
\begin{tabular}{lcc}
\toprule
        & seen items & unseen items \\
\midrule
YCB     & 16         & 26           \\
Conveni & 16         & 24           \\
\bottomrule
\end{tabular}}}
\end{center}
\end{table}

In this study, objects are stacked flat on a table, allowing the robot to approach them from various directions. Across all the experiments, the objects used for evaluation are distinct from those used for training.
Objects of appropriate sizes for grasping by the robot were extracted from the YCB dataset~\cite{YCBDataset}.
To increase the variation in objects, we scanned convenience store items and added scanned 3D models to the set of objects used to create stacked scenes.
To evaluate the generalization performance to unknown objects, we distinguished between "seen objects" and "unseen objects.”
The convenience store items included in the dataset are shown in \figref{fig:conveni_items}, and the number of items in the dataset is listed in Table \ref{tabl:num_3dmodels}.
We used Isaac Sim~\cite{IssacSim} for the simulation.
The number of items for each scene was sampled from a Poisson distribution with a mean $\lambda = 9$. 
Objects with this number were randomly placed to create a piled scene.
A dataset was constructed by recording the contact force and RGB images for each scene. Three images were captured for each of the 2,000 scenes by applying visual domain randomization, including camera parameters, illumination, and object color.
\revadd{We also assumed a constant mass density with object-specific masses, while fixing the friction coefficients (static friction = 0.5, dynamic friction = 0.4).}

\subsection{{\bf RQ1: Does the smoothed distribution make consistent predictions easier compared to the point forces?}}\label{subsec:rq2}

First, we analyze the advantages of smoothing by comparing the prediction results of models trained on data generated with different smoothing methods. To focus on the difference between the smoothing methods, we prepare a baseline using isotropic smoothing with a small variance of $\sigma = 0.005 \mathrm{m}$ as approximated point forces. This value matches the voxel size used to represent the force distribution in 3D space. This baseline is referred to as Isotropic Force Smoothing (IFS($\sigma_f = 0.005$)).
In contrast, a version with stronger smoothing using $\sigma_f = 0.015 \mathrm{m} $ is referred to as IFS($\sigma_f = 0.015$). The proposed method, which applies geometry-aware smoothing, is referred to as Geometry-Aware Force Smoothing (GAFS). We prepare two variants with different smoothing intensities: GAFS($\sigma_f = 0.030$, $\sigma_g = 0.010$) and GAFS($\sigma_f = 0.060$, $\sigma_g = 0.010$).
The only difference among these four methods lies in how the force labels (i.e., prediction targets) are generated; the prediction models and training procedures are identical.
As a comparison measure, we use the force prediction error near the surface of objects, as forces on the surface are what affect object motion.
Furthermore, since the force magnitude can vary widely, errors on heavy objects can be overemphasized.
To mitigate this, we normalize the prediction error by the ground truth force values.
Specifically, we use the following evaluation metrics:
\begin{equation}
L(d) = \Sigma_{{\bf x}\in S(d)}||\hat{y}({\bf x})- y({\bf x})||_2 / \Sigma_{{\bf x}\in S(d)}y({\bf x})
\end{equation}
Here $S(d), \hat{y}({\bf x}), y({\bf x})$ denote a shell of thickness $d$ on the surface of the objects in a scene, predicted force, and ground truth force, respectively.

\subsubsection{Results}

\begin{figure}[tbp]
 \centerline{\includegraphics[width=\columnwidth]{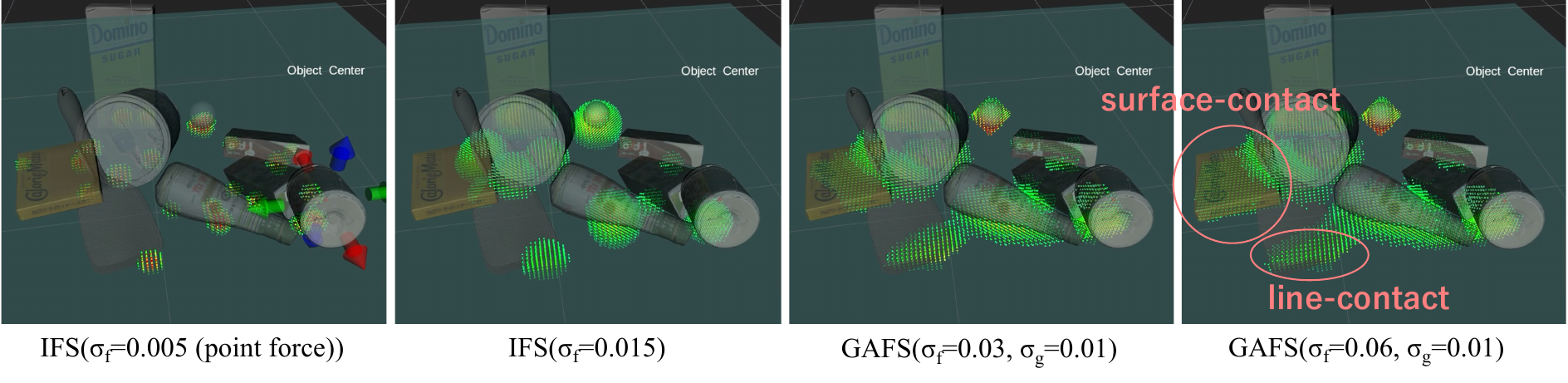}}
\caption{Force distributions generated by different smoothing methods.}
\label{fig:force_label_smoothing}
\end{figure}

\begin{figure}[tbp]
 \centerline{\includegraphics[width=0.7\columnwidth]{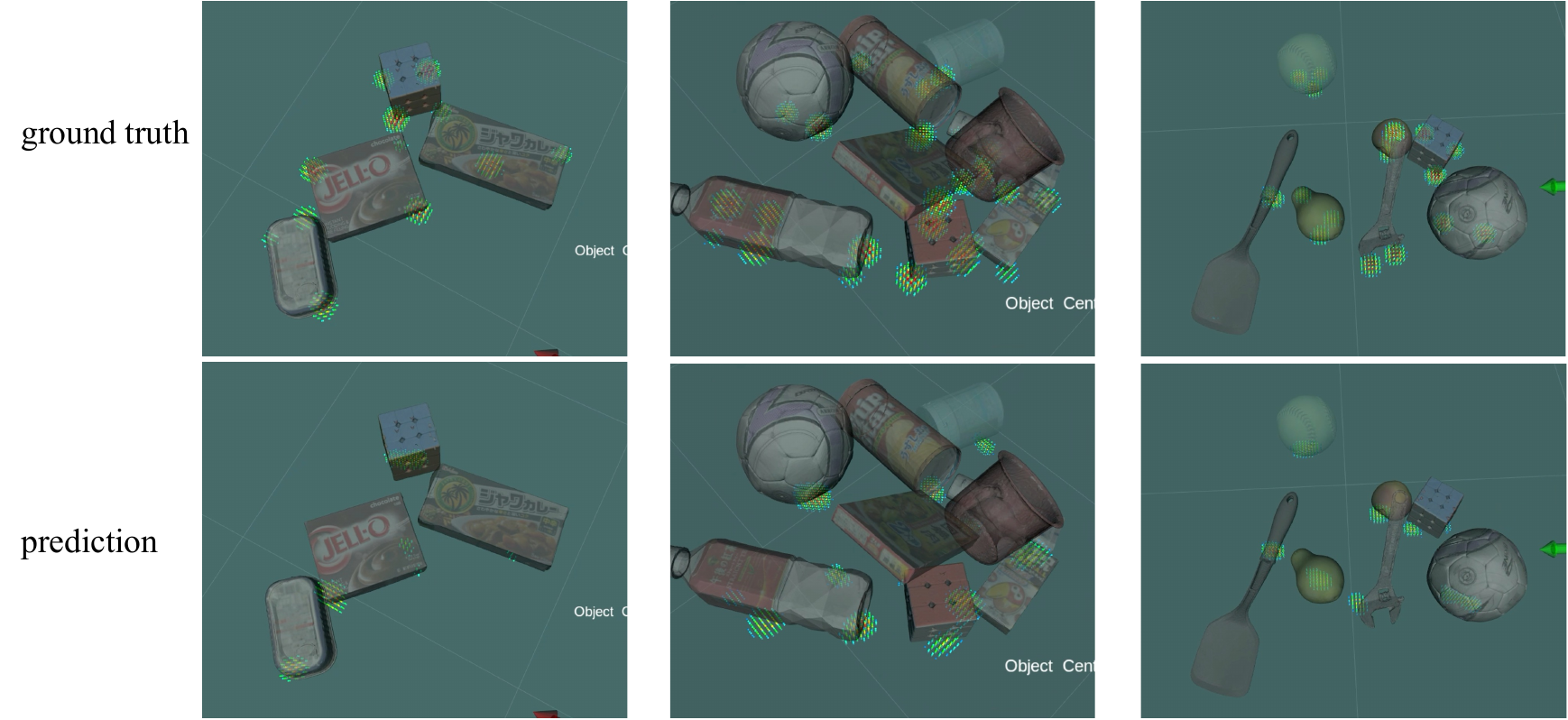}}
\caption{Predicted point forces.}
\label{fig:predicted_pointforces}
\end{figure}

\figref{fig:force_label_smoothing} compares how the smoothed distributions differ among four methods.
The color transition from green to yellow to red indicates increasing force magnitude.
In IFS ($\sigma_f = 0.005$), the forces are concentrated in a small, localized region, whereas with an increased smoothing strength (IFS, $\sigma_f = 0.015$), the distribution becomes more spherically wider.
However, in regions where a box is in surface contact with the table or where the tip of the spatula is in line contact, the distribution only partially covers the contact area.
While increasing the smoothing strength further allows for a broader coverage, the distribution also spreads in the surface normal direction, resulting in a loss of resolution in how the forces are applied.
In contrast, GAFS ($\sigma_f = 0.030$, $\sigma_g = 0.010$) exhibits a distribution more closely aligned with the objects' actual contact surfaces. Line contact areas show elongated patterns, and surface contact regions display flatter patterns. With a higher smoothing strength (GAFS, $\sigma_f = 0.060$, $\sigma_g = 0.010$), the bottom surface of the left box is almost entirely covered by the predicted force distribution.
\figref{fig:predicted_pointforces} shows the predicted force pattern and the ground truth point forces.
Point-like patterns similar to the ground truth are predicted, but some points are missing. In particular, there are several cases where contact points that support the surface are missing, especially in regions where box-like objects are in surface contact with the table.

Table \ref{tab:prediction_errors} shows the prediction errors for different surface thicknesses considered as the object surface. For comparison, the mean squared error (MSE) is also reported. The difference between MSE and $L(d)$ lies in the evaluation domain: MSE assesses the error over the entire voxel grid, while $L(d)$ assesses the error only near the object surface.
For both IFS and GAFS, applying smoothing improves prediction accuracy. Furthermore, comparing GAFS with IFS reveals that incorporating geometry-awareness further reduces the prediction error. This trend becomes more prominent as the evaluation region narrows from MSE to $L(d=1.0)$ and $L(d=0.5)$, i.e., as it becomes more localized to the object surface.

These results suggest that the smoothed distribution is more predictable. However, this does not necessarily mean that it is helpful for manipulation. Therefore, we next investigate the utility of predicting such targets by considering their effectiveness in a downstream task.

\begin{table}[!t]
\centering
\caption{\textbf{Prediction errors near the surface of objects (normalized)}}
\label{tab:prediction_errors}
\setlength{\tabcolsep}{3pt}
\scalebox{0.75}{
\begin{tabular}{ lccc }
\toprule
\textbf{Smoothing Method} & MSE & $L(d=1.0$cm$)$ & $L(d=0.5$cm$)$ \\ 
\midrule
GAFS$(\sigma_f=0.030,\sigma_g=0.010)$ & $0.02548 \pm 0.01017$ & $0.04917 \pm 0.01574$ & $0.05238 \pm 0.01665$ \\
GAFS$(\sigma_f=0.060,\sigma_g=0.010)$ & ${\bf 0.02232 \pm 0.00972}$ & ${\bf 0.04053 \pm 0.01416}$ & ${\bf 0.04240 \pm 0.01484}$ \\
IFS$(\sigma_f=0.015)$ & $0.03139 \pm 0.01149$ & $0.05187 \pm 0.01607$ & $0.05343 \pm 0.01644$ \\
IFS$(\sigma_f=0.005)$ & $0.02308 \pm 0.00619$ & $0.06423 \pm 0.01328$ & $0.08602 \pm 0.01599$ \\
\bottomrule
\end{tabular}}
\end{table}

\subsection{{\bf RQ2: Is the predicted distribution useful for manipulation tasks?}}\label{subsec:lifting_planning}

We consider a downstream task of planning the lifting direction to pick a specified object from a stack, thereby minimizing the disturbance to the surrounding objects as much as possible. Since the algorithm for planning the lifting direction from the predicted forces is the same as in ~\cite{IROS2023:hanai}, its details are provided in the Appendix \ref{sec:lifting_direction_planning}.

The objective here is to evaluate the disturbance caused when lifting is performed using the lifting direction planned from the predicted force distribution. For this evaluation, the following preparation is needed. (1) piled scenes for evaluation, (2) grasp poses with which a robot performs lifting, (3) trajectories for lifting, and (4) metrics to assess the disturbance. (1)-(3) are described in Appendix \ref{sec:lifting_episodes}.

{\bf Evaluation metrics.} There are many options for concrete evaluation measures.
Even if we consider only the change in position, several choices are available, such as the difference between the initial and final positions, the maximum displacement from the initial position during lifting, and the total distance traveled throughout the lifting process.
When several surrounding objects are present, there is also a choice of whether to measure the object with the maximum movement or sum the movements of all objects. 
Velocity, acceleration, contact force, and kinetic energy are also possible evaluation parameters. Among these options, this study adopts the following three evaluation measures.
\begin{itemize}
\item The total distance traveled by objects throughout the lifting process. The distance is summed over all objects except the lifting target.
\item The maximum velocity that occurred during the lifting process. The maximum value is taken for all objects except the lifting target.
\item The maximum contact force that is applied during the lifting process. The maximum value is taken for all objects except the lifting target.
\end{itemize}
The first measure assesses the overall movement of the objects in the scene. We are also interested in reducing the impact on the objects. 
As measures for evaluating the magnitude of this impact, we use the maximum velocity of the object and the maximum contact force acting on it. In this experiment, an object is subjected to a large force when it collides with a table or another object while falling or tipping. Thus, there is a correlation between the object's maximum velocity and the contact force acting on it. However, it is worth noting that the contact force at the moment of impact may vary depending on the object's weight, shape, and other factors.

\revadd{Relating the predicted force distribution directly to downstream task performance is difficult; therefore, our primary evaluation focuses on downstream task performance.
On the other hand, we are also interested in the properties of the predicted distribution itself, which may be related to manipulation tasks. To this end, we introduce additional metrics — Gravity Support (GS) and Torque Balance (TB) - and attempt a physical interpretation.
GS measures how well the predicted contact forces support the object’s weight against gravity. TB measures how well the predicted contact forces balance torques around the center of mass.
However, there are two issues in this evaluation.
First, the predicted force distribution does not specify which object each force acts on. Therefore, we assume that forces near the surface of an object act on that object.
Second, the prediction provides only the magnitude of the force without its direction. To address this, we assume that frictional effects are small and that forces act along the surface normal of the object.
The lifting direction planning algorithm in the Appendix \ref{sec:lifting_direction_planning}
is designed for real-world deployment, where accurate object geometry is not available. Accordingly, the forces acting on the target object are approximated using a sphere, and the force directions are roughly estimated from the gradient of the predicted force. In contrast, the evaluation presented here uses accurate object geometry, resulting in a more precise assessment. Moreover, when deriving force directions from the predicted forces, methods such as IFS (especially with small $\sigma_f$) tend to produce noisy gradients near the object surface. Using force directions derived from the object geometry results in a fairer comparison.}

\revadd{In addition, as a metric for evaluating task relevance, we consider the resistance encountered when lifting the target object in a specified direction. In lifting-direction planning, the relative resistance across candidate lifting directions is important. We evaluate how well the predicted distribution agrees with the ground-truth point-force-based assessment in identifying low-resistance directions. We call this metric Low Resistance Direction Agreement (LRDA).
In lifting planning, the direction with minimum resistance is selected. However, this direction is not always uniquely defined. For example, when an object is placed on a table with another object on top, horizontal lifting directions for the lower object can have equally low resistance. Therefore, instead of focusing on a single optimal direction, we consider all candidate directions and evaluate how well the predicted distribution identifies low-resistance directions overall. This allows us to assess whether the model captures force patterns that are relevant to task performance in a holistic manner.
The detailed computation of each metric is described in the Appendix \ref{sec:app_pi_measures}.}

\subsubsection{Results (lifting in simulation)}\label{sec:results_in_simulation}

\begin{figure}[tbp]
 \centerline{\includegraphics[width=\columnwidth]{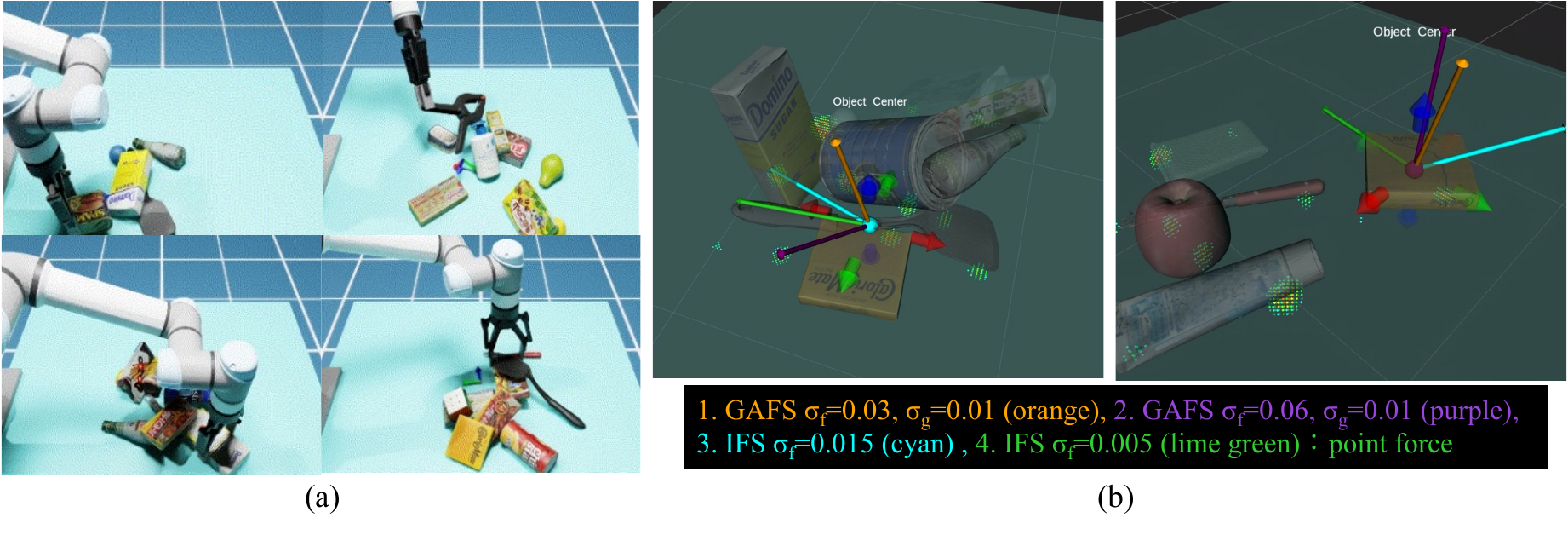}}
\caption{Predicted lifting directions in simulation.}
\label{fig:predicted_directions_in_sim}
\end{figure}

\begin{table}[!t]
\centering
\caption{\textbf{Disturbance caused by lifting}}
\label{tab:disturbance_sim}
\setlength{\tabcolsep}{3pt}
{\scalebox{0.75}{
\begin{tabular}{ lccccc }
\toprule
\textbf{Smoothing Method} & \textbf{Total dist.[m]$\downarrow$} & \textbf{Max vel.[m/s]$\downarrow$} & \textbf{Max force[N]$\downarrow$} & \textbf{Success rate$\uparrow$} \\ 
\midrule
\revadd{No-Force(UP)} & $\bf{0.263 \pm 0.196}$ & $0.741 \pm 0.513$ & $8.877 \pm 16.733$ & $\bf{0.992}$ \\
GAFS$(\sigma_f=0.030,\sigma_g=0.010)$ & $0.349 \pm 0.260$ & $\bf{0.561 \pm 0.371}$ & $\bf{7.654 \pm 7.067}$ & $0.979$ \\
GAFS$(\sigma_f=0.060,\sigma_g=0.010)$ & $0.329 \pm 0.247$ & $0.565 \pm 0.433$ & $8.070 \pm 9.698$ & $0.992$ \\
IFS$(\sigma_f=0.015)$ & $0.324 \pm 0.266$ & $0.572 \pm 0.412$ & $7.930 \pm 10.048$ & $0.996$ \\
IFS$(\sigma_f=0.005)$ & $0.374 \pm 0.278$ & $0.573 \pm 0.566$ & $7.968 \pm 11.027$ & $0.971$ \\
\bottomrule
\end{tabular}
}}
\end{table}

\figref{fig:predicted_directions_in_sim} shows (a) the experiment in progress and (b) the planned lifting directions for two scenes. In the left scene of (b), the model attempts to lift a spatula, but multiple point contacts exist around it. As a result, IFS, which minimizes the work against the point forces arising from those contacts, suggests a direction that intrudes into adjacent objects.
In contrast, GAFS takes into account the object geometry near the contact points and, consequently, finds a direction with less intrusion into neighboring objects, thereby pulling the spatula forward.
A simple case in which the geometry weight shows its effectiveness is shown in the right scene, where a box-shaped object rests on a table. In such cases, multiple point forces typically appear somewhere on the contact surface (empirically, often near the corners), but it is difficult to predict all of them.
When some of these forces are missing from the prediction, the model suggests a diagonal direction, even though lifting straight upward would be appropriate.

Table \ref{tab:disturbance_sim} presents a quantitative comparison. We also include the upward lifting method \revadd{without using force information (No-Force(UP))} as a baseline.
Compared to \revadd{No-Force(UP)}, the other four methods show substantial reductions in both maximum velocity (23–24\%) and maximum contact force (9–14\%). In terms of total distance, the upward lifting strategy was effective; methods that move objects in more horizontal directions tend to risk pushing away surrounding objects, especially in complex scenes with many objects, resulting in worse performance.
We observed improvements by applying smoothing to the force labels. While the improvement in maximum velocity was trivial, a noticeable gain was achieved in maximum force. Most notably, the application of smoothing significantly reduced the variance in both maximum velocity and maximum force, indicating that smoothing helps consistently identify lifting directions that minimize disturbances across diverse scenes. In the comparison of different smoothing methods, GAFS, which takes geometry into account, achieves smaller disturbances than IFS. While Section \ref{subsec:rq2} showed that stronger smoothing in GAFS leads to better prediction accuracy, the results here suggest that excessive smoothing should be avoided in the context of the lifting task performance.

\revadd{Given the effectiveness of geometry-based weights, one may question whether geometry-based cues alone could provide similar information. }
Although Equation \ref{eq:gafs} defines point forces as being weighted by \revadd{the geometry of the colliding objects}\revdel{their distance from the object surface}, it can also be interpreted in reverse—that \revadd{the geometry is weighted by force}\revdel{force predictions apply weights to the geometry}. In other words, increasing the geometry-aware smoothing strength enhances the emphasis on the object’s geometry in the prediction.
\revadd{The observed performance degradation when increasing $\sigma_f$ in GAFS to 0.060 suggests that placing too much weight on geometry weakens the locality of the forces, leading to degraded performance in the lifting task.}
\revdel{One might assume that minimizing disturbances during lifting could be achieved simply by moving objects along the surface normals of contacting objects, without predicting forces. However, these results suggest that explicitly considering force information enables more effective suppression of disturbances than relying solely on surface normals.}
\revadd{In general, while geometry-based methods are suitable to identify collision-free lifting directions, they are inherently limited in scenarios where no such direction exists. In cluttered environments, better lifting often requires controlled interaction with surrounding objects. In such cases, it is important to identify directions that may involve slight contact but result in minimal disturbance.}

Since the evaluation episodes were selected based on successful grasps under the \revadd{No-Force(UP)} policy, a slight drop in task success rate was observed in other methods. Nonetheless, all methods generally succeeded in lifting the object above the required height threshold.

\begin{figure}[tbp]
 \centerline{\includegraphics[width=\columnwidth]{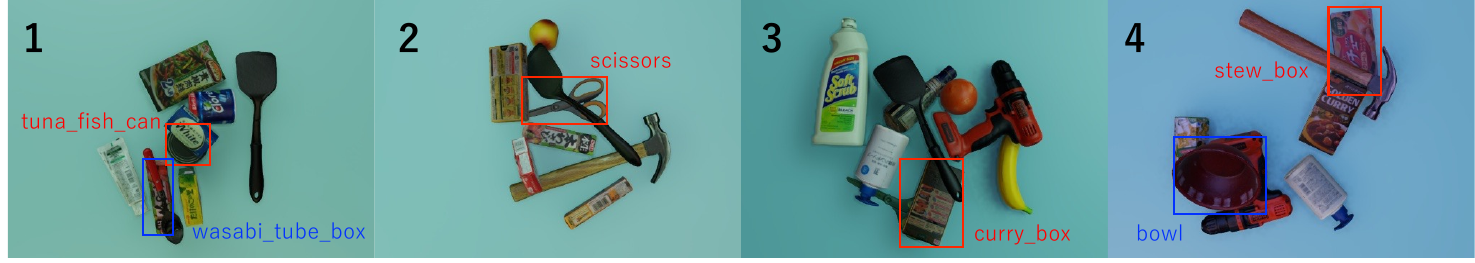}}
\caption{Physical interpretation: scenes for evaluation}
\label{fig:physical_interpretation_scenes}
\end{figure}

\begin{table*}[!t]
\centering
\caption{\textbf{\revadd{Physical interpretation of predicted distribution (GS: Gravity Support, TB: Torque Balance, LRDA: Low Resistance Direction Agreement)}}}
\label{tab:physical_interpretation}
\setlength{\tabcolsep}{3pt}
{\scalebox{0.75}{
\begin{tabular}{ llcccccc }
\toprule
\textbf{Method} 
& \textbf{Metric} 
& \textbf{1,tuna\_fish\_can}
& \textbf{1,wasabi\_tube\_box}
& \textbf{2,scissors}
& \textbf{3,curry\_box}
& \textbf{4,stew\_box}
& \textbf{4,bowl} \\
\midrule
\begin{tabular}[c]{@{}c@{}} \textbf{GAFS} \\ $\sigma_f=0.030$ \\ $\sigma_g=0.01$ \end{tabular} 
& \begin{tabular}[c]{@{}c@{}}GS$\downarrow$ \\ TB$\downarrow$ \\ LRDA$\uparrow$ \end{tabular}
& \begin{tabular}[c]{@{}c@{}}\textbf{0.273} \\ 0.057 \\ \textbf{0.739}\end{tabular}
& \begin{tabular}[c]{@{}c@{}}\textbf{0.486} \\ 0.122 \\ 0.291\end{tabular}
& \begin{tabular}[c]{@{}c@{}}0.428 \\ 0.273 \\ \textbf{0.538}\end{tabular}
& \begin{tabular}[c]{@{}c@{}}0.366 \\ 0.209 \\ 0.379\end{tabular}
& \begin{tabular}[c]{@{}c@{}}1.775 \\ 0.130 \\ 0.429\end{tabular}
& \begin{tabular}[c]{@{}c@{}}0.719 \\ 0.331 \\ 0.500\end{tabular} \\
\midrule
\begin{tabular}[c]{@{}c@{}} \textbf{GAFS} \\ $\sigma_f=0.060$ \\ $\sigma_g=0.01$ \end{tabular} 
& \begin{tabular}[c]{@{}c@{}}GS$\downarrow$ \\ TB$\downarrow$ \\ LRDA$\uparrow$ \end{tabular}
& \begin{tabular}[c]{@{}c@{}}0.350 \\ 0.103 \\ 0.667\end{tabular}
& \begin{tabular}[c]{@{}c@{}}0.487 \\ \textbf{0.096} \\ 0.290\end{tabular}
& \begin{tabular}[c]{@{}c@{}}0.543 \\ \textbf{0.115} \\ 0.333\end{tabular}
& \begin{tabular}[c]{@{}c@{}}0.412 \\ 0.305 \\ 0.333\end{tabular}
& \begin{tabular}[c]{@{}c@{}}\textbf{1.186} \\ \textbf{0.105} \\ \textbf{0.538}\end{tabular}
& \begin{tabular}[c]{@{}c@{}}\textbf{0.708} \\ 0.304 \\ 0.552\end{tabular} \\
\midrule
\begin{tabular}[c]{@{}c@{}} \textbf{IFS} \\ $\sigma_f=0.015$ \end{tabular} 
& \begin{tabular}[c]{@{}c@{}}GS$\downarrow$ \\ TB$\downarrow$ \\ LRDA$\uparrow$ \end{tabular}
& \begin{tabular}[c]{@{}c@{}}0.694 \\ 0.069 \\ 0.600\end{tabular}
& \begin{tabular}[c]{@{}c@{}}0.879 \\ 0.148 \\ 0.290\end{tabular}
& \begin{tabular}[c]{@{}c@{}}0.682 \\ 0.237 \\ 0.481\end{tabular}
& \begin{tabular}[c]{@{}c@{}}0.612 \\ \textbf{0.050} \\ \textbf{0.538}\end{tabular}
& \begin{tabular}[c]{@{}c@{}}1.602 \\ 0.283 \\ 0.429\end{tabular}
& \begin{tabular}[c]{@{}c@{}}0.806 \\ \textbf{0.195} \\ \textbf{0.607}\end{tabular} \\
\midrule
\begin{tabular}[c]{@{}c@{}} \textbf{IFS} \\ $\sigma_f=0.005$ \end{tabular} 
& \begin{tabular}[c]{@{}c@{}}GS$\downarrow$ \\ TB$\downarrow$ \\ LRDA$\uparrow$ \end{tabular}
& \begin{tabular}[c]{@{}c@{}}0.735 \\ \textbf{0.049} \\ 0.538\end{tabular}
& \begin{tabular}[c]{@{}c@{}}0.723 \\ 0.157 \\ \textbf{0.429}\end{tabular}
& \begin{tabular}[c]{@{}c@{}}\textbf{0.307} \\ 0.720 \\ 0.379\end{tabular}
& \begin{tabular}[c]{@{}c@{}}\textbf{0.149} \\ 0.334 \\ 0.143\end{tabular}
& \begin{tabular}[c]{@{}c@{}}4.253 \\ 0.453 \\ 0.379\end{tabular}
& \begin{tabular}[c]{@{}c@{}}0.877 \\ 0.444 \\ 0.324\end{tabular} \\
\bottomrule
\end{tabular}
}}
\end{table*}

\revadd{Table \ref{tab:physical_interpretation} presents a comparison of GS, TB, and LRDA. The scenes and target objects used for evaluation are shown in \figref{fig:physical_interpretation_scenes}.
For GAFS, GS typically ranges from approximately 0.2 to 0.5 across many scenes. This is due to several approximations made in the evaluation. In particular, since the model predicts a smoothed target, some predicted forces fall outside the region considered as the object surface. As a result, the predicted forces tend to be underestimated and cannot fully counteract gravity.
The scene in which the stew\_box target yields GS values greater than 1 is particularly challenging. Since a relatively light object supports a much heavier object placed on top, even small prediction errors can result in large deviations in the evaluation metrics. In addition, the hammer has a significant mass concentrated at its head, and the model has learned that large contact forces are generated when this part contacts the table. As a result, the model predicts contacts at locations that are significantly offset from the actual contact points. In our evaluation, this leads to incorrect large horizontal force components that are not canceled by gravity.
In terms of comparison across methods, GAFS generally produces smaller values. While TB is sometimes better for IFS, this is mainly because the prediction error of forces arising from contacts with upper objects is large, resulting in predictions that resemble a scenario where the object is simply placed on the table. Consequently, GS also becomes larger in such cases.
LRDA is significantly improved by smoothing. Comparing GAFS and IFS, GAFS achieves higher values in cases where an object is supported by surface contact on a tuna\_fish\_can, or target object is flat like scissors. In contrast, IFS performs better in cases such as the curry\_box or bowl, where accurate localization of point contacts is critical.
IFS with weak smoothing generally yields poorer performance across all metrics. Although it occasionally produces high values in some scenes, the results are highly variable, indicating that it does not provide consistent predictions across diverse scenes.}

\subsection{{\bf RQ3: Does the prediction model transfer well to real-world scenes?}}

\begin{figure}[tbp]
 \centerline{\includegraphics[width=0.7\columnwidth]{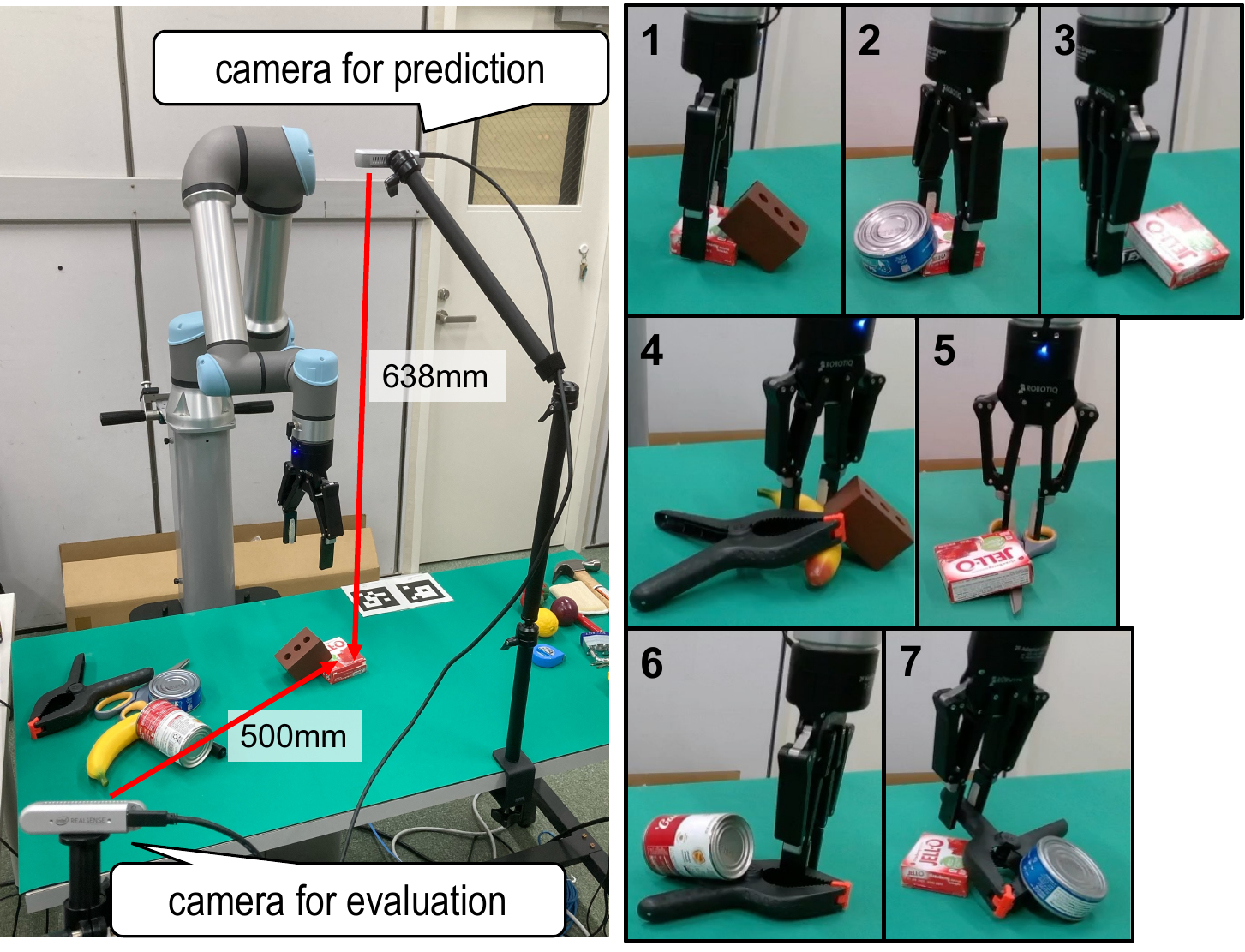}}
\caption{Evaluation system and scenes: top-view camera is used for prediction and front camera is used for evaluation. Note that these cameras are RealSense; however, depth is not used for force prediction. 
The point cloud from the top-view camera is used solely for visualizing the positional relations between objects and the predicted force. The right figures are scenes used for evaluation.}
\label{fig:evaluation_system_and_scenes}
\end{figure}

One of our major interests lies in whether a model trained solely in simulation can make useful predictions for real-world scenes, just as it does in simulation. To investigate this, we utilize a model trained only in simulation to predict force distributions in real piled scenes and evaluate its performance. However, since it is difficult to measure the forces acting on objects in a piled scene without affecting the scene in a real environment, the evaluation is conducted through a qualitative assessment of the predicted force distribution and a quantitative evaluation based on the performance of the lifting task.

\subsubsection{Object tracking in real cluttered scenes for evaluation}\label{subsec:tracking}

In a real environment, it is not possible to directly obtain the forces acting on objects or their positions as can be done in a simulator. Therefore, we use vision to track the motion of surrounding objects and evaluate disturbances based on their trajectories.
For this tracking, we used FoundationPose~\cite{Wen2023FoundationPoseU6}.
FoundationPose takes RGBD measurements and a 3D model of an object, and estimates the 6DoF pose of the object.
We used RealSense D415 for the RGBD camera.
As FoundationPose requires a segmentation mask for the object to initialize the tracker,
we used YOLO v5~\cite{ultralytics2021yolov5} for the instance segmentation from an image.
A YOLOv5s model was trained using the YCB video dataset~\cite{Xiang2017PoseCNNAC}
following the PoseCNN training procedure~\cite{Xiang2017PoseCNNAC}. 
Real and synthetic data were used to train the instance segmentation model.

\subsubsection{Evaluation system and protocol}

The setup for the real-world experiment is shown in \figref{fig:evaluation_system_and_scenes},
with a camera for prediction directly above the table and a camera for evaluation in front of it. The robot used is UR5e and the gripper is Robotiq 2F-140.
To fit the fingers to concave surfaces and complex shapes, a gel measuring 1 cm in width and 0.5 mm in thickness, which is narrower than a finger width, was attached to the fingers.
The lifting motion was designed to pull out the target object by 15 cm along the direction planned by each method, and the gripping force was set to 62.5N, which was 50\% of the maximum gripping force of 2F-140. YCB objects are typically empty. To reproduce realistic weights, we applied weights as close as possible to the original weights of the boxes and cans.

We prepared seven piled scenes for evaluation (\figref{fig:evaluation_system_and_scenes}).
In each scene, the object to be picked was placed in a graspable arrangement, and
a grasp pose was defined.
For each scene, we performed three lifting trials 
\revadd{by creating the same piled configuration at three locations: 5 cm from the predefined center point toward the front, back-right, and back-left. Then, we evaluated the disturbance by analyzing the trajectories of surrounding objects.}
\begin{figure}[tbp]
 \centerline{\includegraphics[width=0.5\columnwidth]{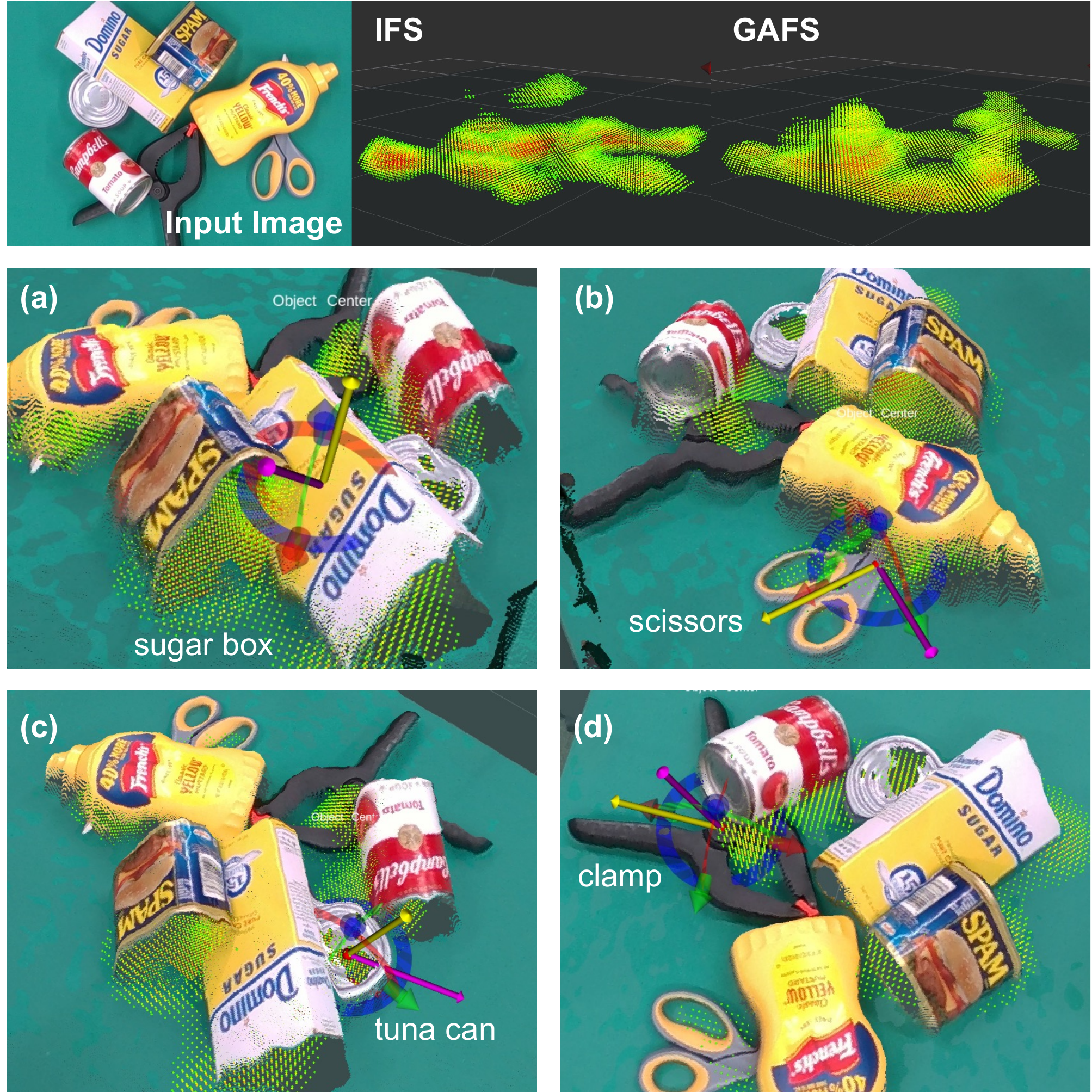}}
\caption{Top-right images show the predicted force patterns. As the value becomes large, the color changes from green to red. Small forces are not drawn for ease of viewing.
Planned lifting directions computed using the proposed method (GAFS) and a baseline (IFS) are shown in the bottom row. The magenta arrows show the directions obtained from IFS, and the yellow arrows show the directions obtained from GAFS． GAFS worked better for most of the objects.}
\label{fig:planned_lifting_directions}
\end{figure}

{\bf Evaluation metrics and baselines.} The impulsive force acting on an object can be calculated from the change in momentum of the object. However, the calculation requires the acceleration of the object, and evaluating acceleration is noisy. Therefore, for the evaluation in the real scenes, we only use the maximum velocity to assess the impact on the objects.
These measures are evaluated separately for translation and rotation, resulting in four distinct measures. Through the experiment, we confirmed that FoundationPose tracked the poses of surrounding objects surprisingly well. One reason is that the depth of RealSense D415 is relatively stable, even in fast-moving scenes.
Although the scenes for evaluation included certain occlusions, there were no situations in which the surrounding objects were largely hidden by occlusion.
However, one problem was identified in the evaluation of the rotation. 
Stable tracking was difficult for rotationally symmetric objects, such as cylindrical cans, despite their textures.
Sometimes, the tracker failed to follow the posture change of the rolling can.
Fine posture oscillations were sometimes observed in a stationary can.
The former leads to underestimation of the amount of movement, whereas the latter leads to overestimation. Neither of these evaluations is valid.
Therefore, we evaluated the rotation and rotational velocity using only the rotations around the two axes, excluding those around the axis of symmetry.
For GAFS and IFS, we use the parameters that showed better performance in the simulation experiments—GAFS($\sigma_f=0.030$, $\sigma_g=0.010$) and IFS($\sigma_f=0.015$)—and refer to them simply as GAFS and IFS, respectively.

\subsubsection{Results (lifting in the real-world)}

\figref{fig:planned_lifting_directions} shows the lifting directions computed from IFS and GAFS. All objects in this scene are unseen objects that are not included in the training data.
The upper-left corner of the figure shows the input image of the target scene.
The upper right panel shows the predicted force distributions.
Both IFS and GAFS can predict three-dimensional distributions. 
They predict forces not only at the points of contact with the table surface
but also at the points of contact between objects.
The IFS predicts a more point-like pattern, whereas GAFS predicts a smoother pattern for real scenes considering the geometry, as expected.
Fig. (a)–(d) show the planned lifting directions for each object.
The magenta arrows represent the lifting directions obtained using the IFS,
and the yellow arrows represent the lifting directions obtained by GAFS.
In (a), the IFS indicates the direction wherein the SPAM can be pushed away, 
whereas GAFS exhibits the direction with respect to the surface constraint between the SPAM can and the sugar box.
(b) shows the results of specifying a flat object, scissors, which is one of the cases wherein the IFS does not work very well. GAFS exhibits a more horizontal direction.
In (c), a tuna can is sandwiched between a sugar box and the SPAM.
Both IFS and GAFS are appropriate because they yield the directions between these two surrounding objects.
GAFS appears better because it directs diagonally upward to avoid pushing the tomato can.
In (d), a black clamp is specified to pick. A heavy can of tomato soup is located on top of the clamp. 
GAFS indicates a more horizontal direction than the IFS.
In scenes (a), (c), and (d), GAFS functions better for multi-contact situations.

\begin{figure}[tbp]
 \centering\includegraphics[width=\textwidth]{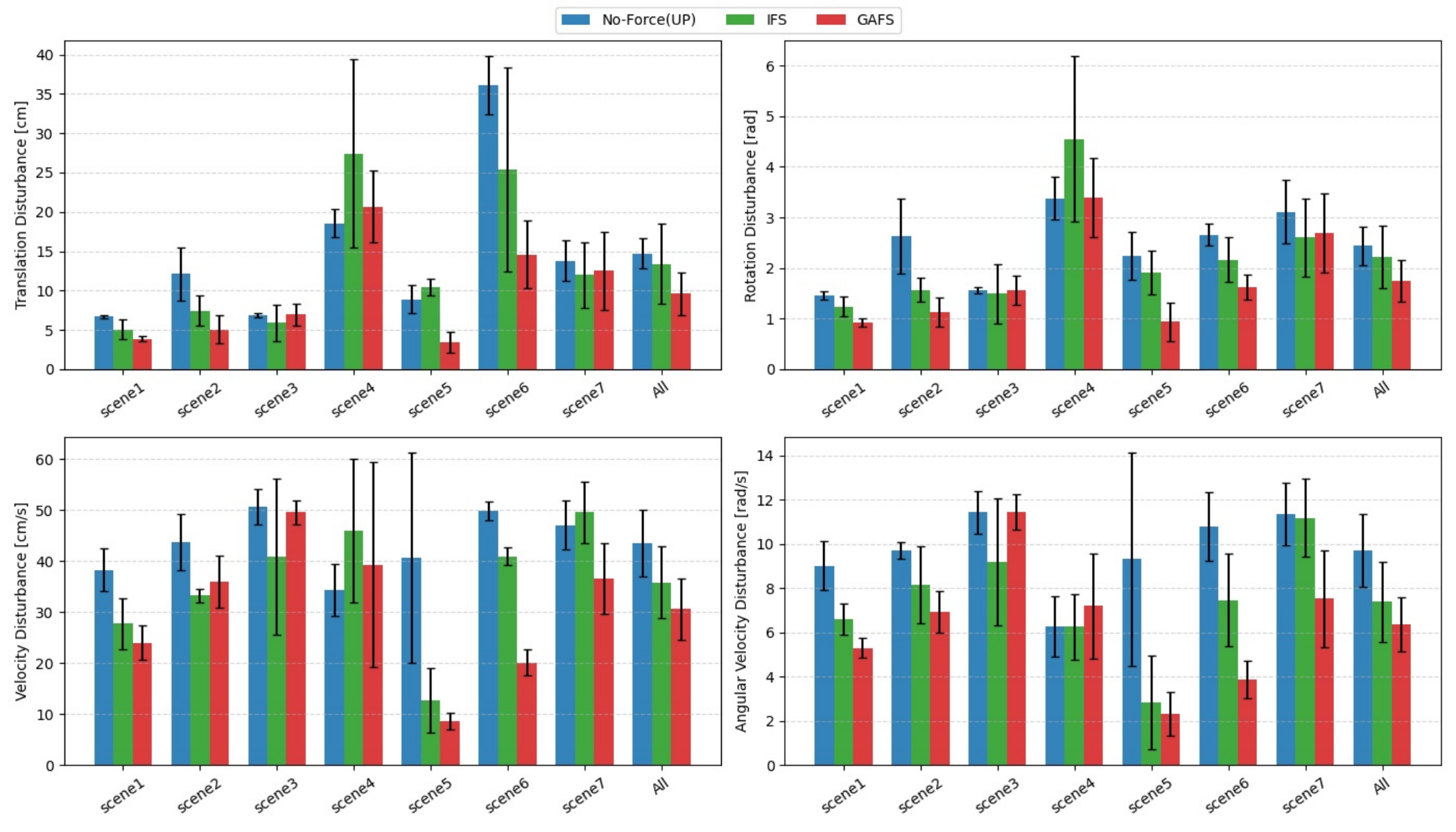}
\caption{Disturbance to the surrounding objects while lifting.}
\label{fig:lifting_results}
\end{figure}

Figure \ref{fig:lifting_results} compares the disturbances during lifting. In most scenes, GAFS achieved the smallest disturbances. In \ref{sec:results_in_simulation}, the total distance of GAFS was larger than that of \revadd{No-Force(UP)}. This is because, in the complex scenes of \ref{sec:results_in_simulation}, interactions occur with many objects. In particular, when movements that push aside multiple objects arise, the total distance increases significantly. In contrast, in the experiments of this section, only a small number of objects interact with the lifting target, and the way of interaction can be predicted before lifting, resulting in the higher performance of GAFS.
As for the standard deviation, the \revadd{No-Force(UP)} always lifted in the same direction, resulting in a small value. The overall standard deviation of GAFS was as small as that of the \revadd{No-Force(UP)} and smaller than that of IFS. This indicated that GAFS found the directions contributing to smaller disturbances more consistently than the IFS.
In Scene 3, the robot picked a marker hidden in a box.
We found that when the target was mostly hidden or small, both IFS and GAFS were less effective.
In this scene, all three methods output almost the same direction (vertically upward), and the difference was small.
For all the evaluation measures, the overall trend was consistent. GAFS achieved the smallest disturbance in most scenes. The detailed analysis of each scene is provided in the Appendix \ref{sec:app_real_lifting}.

\section{\revadd{Limitations and Future Extensions}}

\revadd{We summarize limitations and potential future extensions of the proposed approach.}

\subsection{\revadd{Uncertainty in simulation-generated forces}}

\revadd{Since the supervision signal in the proposed method is generated by a rigid-body simulator, the resulting contact forces depend on simulator settings such as mesh resolution, contact solver parameters, and discretization. Furthermore, in multi-contact systems, even when the contact points are identical, there remains freedom in how forces are distributed among them. As a result, contact force distributions are generally not uniquely determined, and different simulator algorithms or numerical processes may produce different outputs.}
\revadd{Therefore, the proposed method does not aim to accurately reproduce a unique simulator output. Instead, the learning target should be interpreted as one plausible simulator-consistent solution. To reduce variability caused by simulator settings and numerical processes, we apply statistical smoothing. For example, a highly localized distribution of forces among contact points is transformed into smoother distributions over contact regions.}

\subsection{\revadd{Limitations arising from the physical modeling gap}}

\revadd{The ability to handle different physical properties is limited. While differences in mass are modeled, we assume uniform mass density and fixed friction coefficients. As a result, prediction accuracy may degrade in cases where the center of mass is significantly offset or where friction differs from the assumed values. As future work, this limitation could be addressed by using datasets that include object-specific physical properties (e.g., center of mass and friction coefficients), or by leveraging LLM/VLM-based approaches to estimate coarse physical properties for large-scale 3D object datasets.}
\revadd{Since we assume rigid bodies, the method cannot handle large deformations. However, small deformations at the surface are implicitly handled by smoothing, which treats force estimation as a rough prediction problem and reduces sensitivity to differences in stiffness.}
\revadd{Regarding internal contents, the model is trained assuming objects are filled. For properties that cannot be inferred from visual input alone, extensions such as learning multiple hypotheses and selecting among them using additional modalities such as tactile/force sensing, or contextual information would be necessary. 
}

\subsection{\revadd{Limitations arising from smoothing}}

\revadd{When smoothing is too strong, the predicted force distribution may spread into regions belonging to mechanically independent objects. Since the model does not explicitly associate predicted forces with specific objects, this may negatively affect downstream tasks. For example, in lifting direction planning, forces originating from neighboring contacts may be incorrectly included, particularly in scenes with thin stacked objects.}
\revadd{Compared to IFS, GAFS mitigates this issue by incorporating object geometry during smoothing, which suppresses the spread of force distributions beyond contacting objects.}
\revadd{This issue is closely related to the smoothing strength. In our experiments, we set $\sigma_g$ = 0.010 (1 cm), which is smaller than $\sigma_f$. This choice reflects the assumption that distributed contacts occur along surfaces or lines, and that modeling errors in the normal direction and deformation are relatively small. On the other hand, if $\sigma_g$ is too small, the expressiveness of the distribution decreases and may negatively affect lifting direction planning based on gradients. Therefore, we set $\sigma_g$ to approximately twice the voxel size. While more optimal values may exist depending on the application, we observed improved performance with this setting. Reducing $\sigma_g$ can also help mitigate the aforementioned over-smoothing effects.}

\subsection{\revadd{Factors affecting sim-to-real transfer}}

\revadd{One reason for successful sim-to-real transfer in our experiments is that the physical conditions do not significantly deviate from the assumptions described above. In addition, geometry-aware smoothing reduces sensitivity to simulation conditions and contact configurations, which contributes to consistent prediction.}
\revadd{Though we use fixed values for many physical parameters, randomization on these physical parameters may reduce overfitting to specific similator settings and further improve transfer performance to the real world.}
\revadd{Regarding the sim-to-real gap in input images, simple domain randomization worked better than expected. While real images tend to produce slightly blurrier distributions, qualitatively similar patterns are still obtained. Using more powerful vision encoders may further improve the ability to recognize unseen objects and thus enhance force prediction performance.}
\revadd{Camera parameters (extrinsics and intrinsics) are slightly randomized during training. Therefore, precise alignment of camera positions is not required, although large deviations can degrade prediction accuracy.}
\revadd{The prediction performance also decreases for objects that are difficult to perceive visually. For example, while contact between transparent objects and the table can be estimated to some extent, predicting contacts between transparent objects and other objects remains challenging.}

\section{CONCLUSIONS}

To investigate the hypothesis that predicting a rough contact force distribution from the visual input is useful for object handling, we proposed a method for predicting the contact forces acting on objects in a piled scene.
This method uses a rigid-body simulator, which is widely used in robotics. To reduce the gap between the point forces of the simulator and the real-world forces, it applies statistical smoothing to the force label.
Through extensive evaluation, we confirmed that the proposed smoothing yields consistent results in both the predicted forces themselves and in downstream task performances. We also proposed a geometry-guided smoothing method and demonstrated its superiority over isotropic smoothing. Moreover, we verified that the trained model can be transferred to real-world environments, despite being trained solely on synthetic data. 

\revdel{Various extensions are possible for future work.
Conditioning with additional contexts such as languages and tactile/force sensing, as mentioned in the {\it Introduction}, is one such example.}
\revadd{We also discussed the limitations and potential extensions of the proposed approach from multiple perspectives.} In this work, predictions were made prior to lifting. \revdel{It would be possible to predict} \revadd{Predicting} forces continuously while the robot is lifting, 
\revadd{enabling the robot to update its motion in response}
\revdel{such that the lifting motions are updated according} to the changes in the contact state \revadd{would be an interesting direction}.

\section*{Acknowledgement(s)}

We would like to thank Naoya Chiba of Osaka University for the useful discussions.
This work was supported by JST [Moonshot R\&D][Grant Number JPMJMS2031].
During the preparation of this manuscript, the authors used GPT5.3(ChatGPT) for the translation of some statements and grammar checking. After using this, the authors reviewed and edited the content as needed and take full responsibility for the content of the publication.

\bibliographystyle{IEEEtran}
\bibliography{force_estimation}


\appendix

\section{Implementation Details of Force Distribution Prediction and Lifting Direction Planning}

\begin{table}[tbp]
\caption{Parameter values used for the experiments}
\begin{center}
\begin{tabular}{c|c}
\hline
Property          & Value \\ \hline\hline
Force distribution grid size & 0.5 cm \\
Force label smoothing strength (IFS) & $\sigma_f$=1.5 mm \\
Force label smoothing strength (GAFS) & $\sigma_f$=3.0 cm \\
Geometry-based force label smoothing strength & $\sigma_g$=1.0 cm \\
Target object radius & 5.0 cm \\
LeakyReLU slope for lifting direction planning & 0.05 \\ \hline
\end{tabular}
\label{tabl:parameters}
\end{center}
\end{table}
The parameters used for the experiments are summarized in Table \ref{tabl:parameters}.
A small voxel grid size enables high resolution. However, it leads to a larger model size and longer training time. Therefore, a practical grid size of 0.5 cm was chosen, which is slightly smaller than the thickness of the scissors' handle used in the experiments.
Since the force distribution in real environments is unknown and the concept of our approach is to predict rough patterns that humans are likely to imagine, the parameters for force smoothing were determined empirically by observing smoothed patterns across several scenes.
The smoothing strength of GAFS was set to decay gradually compared to the above grid size. Actually, the smoothing parameter for GAFS does not require precise tuning, as even a larger value the smoothed force decays away from the contact surface due to the influence of geometry-based weights.
On the other hand, more caution is required to adjust the force smoothing strength for IFS than for GAFS. This difficulty serves as a motivation for GAFS. Increasing the smoothing strength to cover a broader area reduces the ability to recognize fine geometries. In the experiments, the smoothing strength for IFS was set smaller than that for GAFS.

The radius used to calculate the lifting direction was 5 cm. This choice is related to the size and complexity of the shapes of the objects. Since the force is predicted as a smooth distribution, the lifting direction is not sensitive to changes in this radius. However, the accuracy of the lifting direction decreases if contact with two objects other than the lifting target occurs within this radius or if the lifting target makes contact outside the specified radius. To further improve performance, it is necessary to predict the spread of the target object, which remains one of our future works.

\section{Visual Domain Randomization}

\begin{table}[tbp]
\caption{Visual domain randomization properties and their ranges.}
\begin{center}
\begin{tabular}{c|c}
\hline
Property          & Range \\ \hline\hline
Lighting position [m] & $(0, 0, 2.5) \pm (2.5, 2.5, 1.5)$ \\
Light intensity [lm] & $8500 \pm 7500$ \\
Lighting color [RGB] & $(0.5, 0.5, 0.5) \pm (0.3, 0.3, 0.3)$ \\ \hline
Camera focal length [mm] & $1.8800 \pm 0.0564.0$ \\
Camera horizontal aperture & $2.6034 \pm 0.0781$ \\
Camera vertical aperture & $1.4621 \pm 0.0439$ \\
Camera position [m] & $(0, 0, 1.6)\pm(0.02, 0.02, 0.02)$ \\
Camera orientation [RPY, deg] & $(0, 90, 180)\pm(2, 2, 2)$ \\ \hline
Object diffuse color [RGB] & DefaultColor $\cdot [0.75 \pm 0.25]$ \\
Object specular color [RGB] & $[(0,0,0), (1,1,1)]$ \\ \hline
\end{tabular}
\label{tabl:domain_randomization}
\end{center}
\end{table}
We used visual domain randomization to transfer a simulation-trained model to a real-world environment. Table \ref{tabl:domain_randomization} shows the randomized properties and their respective ranges. For lighting, three sphere lights were placed, and their positions, intensities, and colors were randomized. For the camera, both intrinsic and extrinsic parameters were randomized, and for each object, the color of the material was randomized.
Furthermore, data augmentation was applied to domain-randomized images during training. The data augmentation introduced perturbations in brightness, contrast, and hue for color, as well as translation, rotation, and zoom for geometry and Gaussian noise.

\section{Generation of Lifting Episodes}\label{sec:lifting_episodes}

To minimize arbitrariness, candidate piled scenes for evaluation are generated through random piling, similar to the training data. However, since many of these scenes involve little object overlap and are therefore unsuitable for evaluation, we extract scenes in which (1) a graspable lifting target exists and (2) other objects overlap the lifting target. 

Grasp candidates are generated using the grasp sampler from MIMO~\cite{Cai2024VisualIL}. From an object mesh and a gripper model, grasp candidates are generated and applied to the target scene in Isaac Sim. If the object can be lifted straight upward, the grasp is considered successful, and the scene and grasp pose pair are defined as one evaluation episode. Up to three grasps are generated per lifting target. For difficult scenes in which fewer than three grasps are found among 50,000 candidates, only the available ones are adopted.
However, some of these episodes involve stable stacks where both the lifting target and an object on top are box-shaped and well-aligned. Since the compared methods do not change the object orientation during lifting, such cases often result in the entire stack being transported, regardless of the method. These cases obscure numerical performance differences and are excluded from evaluation. After this filtering process, a total of 96 scenes and 249 episodes were generated.

The lifting motion is designed as follows: The object is first pulled out by 10cm in the direction planned by each method. When this distance is large, the robot arm is more likely to reach its range-of-motion limit, depending on the object placement. Accordingly, 10cm was adopted. If the robot reaches its range limit with only a short pull in a near-horizontal direction, the disturbance can be small. However, this does not reflect a successful lifting of the target object. Therefore, if the object’s height is less than 10cm after pulling, the object is lifted vertically until it reaches a height of 10cm or the arm's limit.

\section{Lifting Direction Planning}\label{sec:lifting_direction_planning}

\begin{figure}[tbp]
\centerline{\includegraphics[width=0.8\columnwidth]{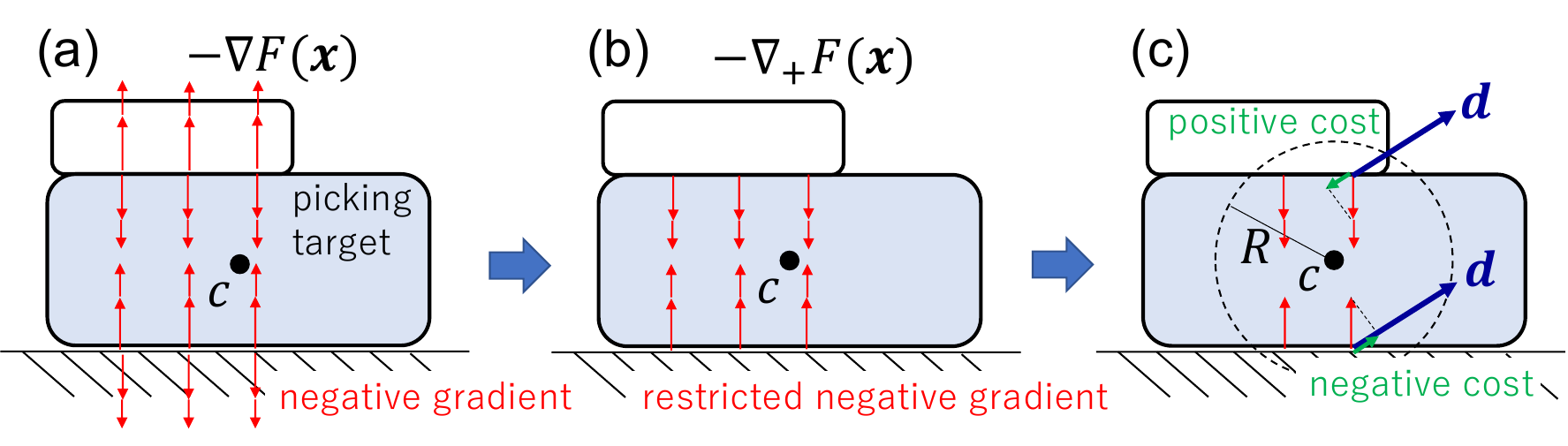}}
\caption{Lifting direction planning algorithm. (a) Force distribution takes large values on the contact surface.
(b) Force distribution gradient is restricted to those with an outward component from $\bf{c}$. Subsequently, the negative value of the restricted gradient approximates the force acting on the target object.
(c) Two costs are computed inside a sphere of radius $R$: cost for penetration and negative cost for repulsion.
}
\label{fig:picking_direction_method}
\end{figure}

The algorithm used to compute the lifting direction from the predicted force distributions is the same as that used in our previous work. Therefore, this concept is briefly described as self-completeness.

\figref{fig:picking_direction_method} illustrates the image of the algorithm.
We assume that the center ${\bf c}$ of the object to be picked $o$ and its approximate size $R$ are given.
Because the predicted force distribution $F({\bf x})$ contains no information on the direction of the force, we first consider the gradient of the force distribution (a).
Then, we restrict the gradients to those with an outward component from ${\bf c}$.
as follows.
\begin{equation}
  \nabla_{+}F({\bf x}) = \left\{ \begin{array}{l}
          \nabla F({\bf x}) \;\;\; {\rm if} \; \nabla F({\bf x})\cdot ({\bf x}-{\bf c}) \geq 0 \\
          {\bf 0} \;\;\; {\rm otherwise}
         \end{array} \right.
\end{equation}
We consider the negative value of $\nabla_{+}F({\bf x})$ to be a virtual force 
which approximates the force acting on the target object.
To further exclude the forces generated at the boundaries between objects other than ${o}$, we limit the evaluation of the force distribution gradient to the interior of 
a sphere with center ${\bf c}$ and radius $R$.
The lifting direction vector ${\bf d}$ is computed to minimize the following equation:
\begin{eqnarray}
  \text{min} && \int_{V({\bf x})}\text{LeakyReLU}(-\nabla_{+}F({\bf x})\cdot {\bf d}) dV \label{eq:lifting_direction_cost} \\
  \text{subject to} && \parallel {\bf d} \parallel = 1
\end{eqnarray}
Equation (\ref{eq:lifting_direction_cost}) considers the following two costs.
When an object is moved along the direction ${\bf d}$, the component against the gradient indicates that it pushes the contacting objects.
Because this should be avoided to the best extent possible, this component incurs a high cost. In addition, if the surrounding objects need not be pushed away, the object is expected to move away from the surface it is in contact with.
For example, if an object is placed on the floor, any movement in any direction, except for the floor direction component, will not apply force to the surrounding objects.
In this case, the object is expected to move away from the floor.
Therefore, we add a small negative cost to the component along the gradient when the object moves in direction ${\bf d}$.
These two costs are combined in (\ref{eq:lifting_direction_cost}).
The negative region of the (LeakyReLU) corresponds to a negative cost. 
The slope of the line in the negative region indicates the weight of negative cost.
In the following experiments, the object radius is set to $R=10$ cm, ensuring it approximately covers most of the objects targeted for grasping.

\section{Detailed analysis of lifting in the real world}\label{sec:app_real_lifting}

\begin{figure}[tbp]
 \centering\includegraphics[width=\textwidth]{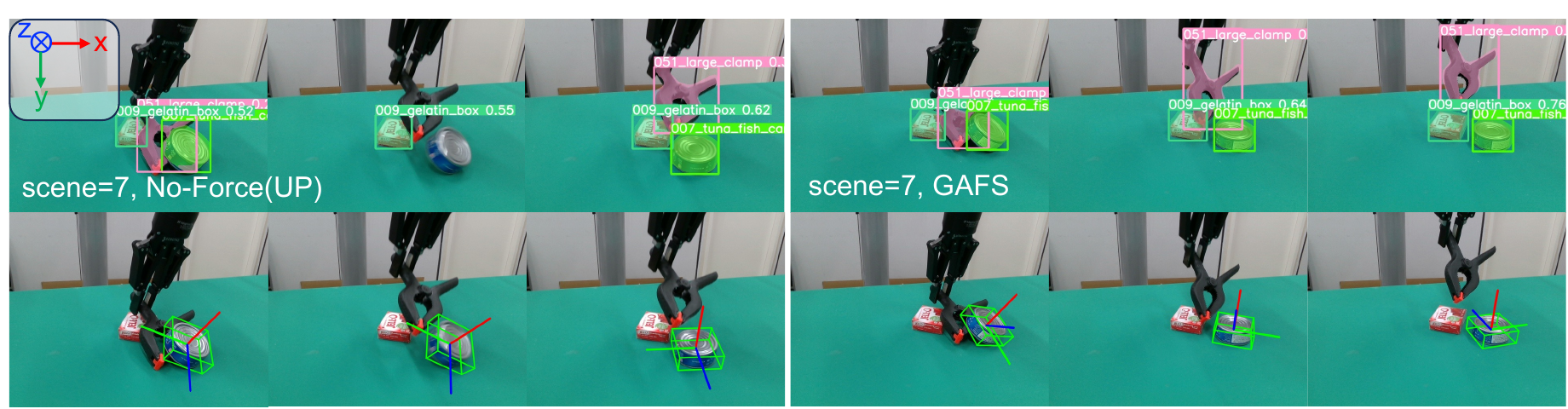}
\caption{Tracking results. The surrounding objects were first detected using YOLOv5 and then tracked with FoundationPose. At the time of evaluation, each object is tracked individually. The tuna can is tracked in this figure.}
\label{fig:tracking_results}
\end{figure}
\begin{figure}[tbp]
 \centering\includegraphics[width=0.7\columnwidth]{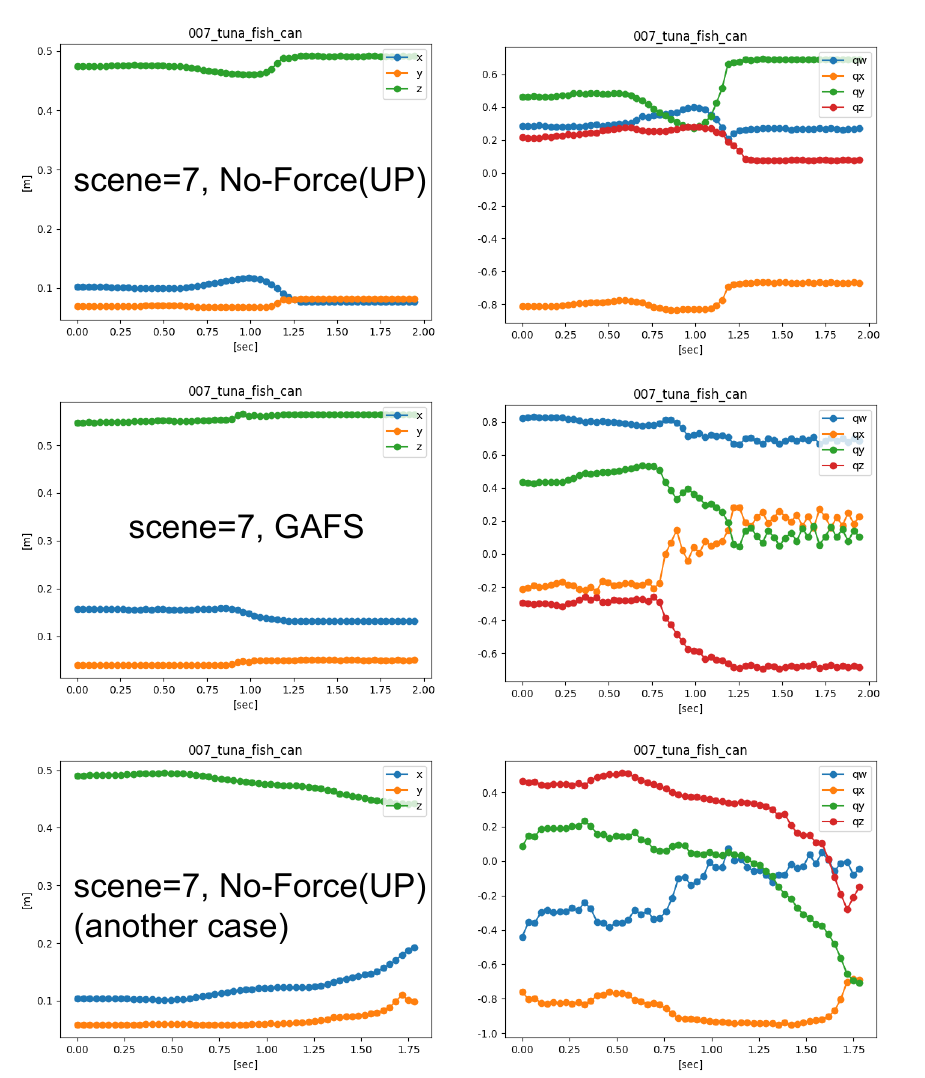}
\caption{Estimated trajectories by FoundationPose. The left column shows the xyz positions of the tuna fish can, and the right column shows its orientations.}
\label{fig:trajectories}
\end{figure}

\figref{fig:tracking_results} shows the results of tracking the surrounding objects. The object poses
are estimated using camera coordinates. In “scene=7, \revadd{No-Force (UP)},” when the robot lifts the
clamp straight up, the reaction of the tuna can was excessively large; consequently,
a sequence of frames of a certain length was not detected by YOLO. When GAFS
was applied, YOLO successfully detected almost all frames, indicating that GAFS
could lift the surrounding objects without causing large velocities. For both sequences,
FoundationPose successfully tracked the tuna can by initializing the pose using the
frame detected by YOLO, as shown in the bottom row of the figure.

\figref{fig:trajectories} shows the estimated trajectories by FoundationPose.
In the first row, the reaction was clean. The tuna fell straight and remained stationary. Thus, the velocity of the can reached the maximum value immediately before colliding with the table.
In the second row, the movement was smaller than that in the first row. However, transient behavior was complex. The can rotated slightly along the round edge and fall. Subsequently, it was slightly bounded. An oscillation was observed in the estimated orientation around the axis of symmetry after the can came to rest. 
The degree of rotation around the axis of symmetry was excluded from the evaluation as described below.
Different types of reactions were observed in the same scene.
In the third row, because the tuna fish can was lifted and overturned,
the translation and rotation movements were large and continued over a long period.

In scene 2, IFS lifted the tuna can higher than GAFS (the translation distance was larger).
However, the translation velocity was smaller because the can dropped by rolling on the round edge. This is confirmed by the large rotational distance and velocity.
In scene 4, other objects covered the target banana from both sides.
Here, \revadd{No-Force (UP)} performed well because the object could be pushed away to both sides by lifting the banana straight up, and this plastic banana was also slippery.
When lifting the banana in a near-horizontal direction, there were cases wherein the clamp was caught at the tip and lifted significantly.
The standard deviations were large in the IFS and GAFS because there were scenes wherein the clamp was caught.
Scene 5 shows the task of pulling scissors from under the box.
Here, \revadd{No-Force (UP)} exhibited significantly higher translation and rotation velocities than the other methods because the box was lifted significantly and fell down.
In one of the three trials, the scissors were dropped because of the large force applied to them.
In contrast, the IFS attempted to pull the box horizontally; however, the direction of the pull was not good. The scissors moved along with the box, resulting in low velocities and large distances.
In scene 6, the translation distance was larger because the tomato could roll after the clamp was removed. The rotation distance was small because the rotation around the axis of symmetry was excluded.
In Scene 7, GAFS attempted to pull out the clamp in a direction closer to the horizontal, which caused the JELLO box underneath it to drag in a trial. 
Consequently, the GAFS lost to the IFS with respect to the translation distance. However, GAFS worked best in terms of translation and rotation velocities.

\section{Metrics for Physical Interpretation}\label{sec:app_pi_measures}

\revadd{
Since the predicted distribution only provides the magnitude of forces, it is necessary to reconstruct the forces acting on each object as well as their directions. To this end, we assume that forces located near an object surface act on that object. Specifically, we define a surface region of thickness $\delta$ around the target object, denoted as $\mathcal{S}_\delta$, and regard the forces within this region as acting on the object. The direction of each force is assumed to align with the surface normal of the object.}
\revadd{
\begin{equation}
\mathbf{F}_i = -f_i \, \mathbf{n}_i, \quad \text{for } i \in \mathcal{S}_\delta
\end{equation}
}

\revadd{Using these reconstructed forces $\mathbf{F}_i$, we define GS and TB as follows. Here, GS is normalized by the magnitude of the gravitational force acting on the object, and TB is normalized by the sum of the magnitudes of the torques acting on the object.}
\revadd{
\begin{eqnarray}
\mathrm{GS} &=& \frac{\left\| \sum_{i} \mathbf{F}_i + m \mathbf{g} \right\|}{m g}, \\
\mathrm{TB} &=& \frac{\left\| \sum_{i} \mathbf{r}_i \times \mathbf{F}_i \right\|}{\sum_{i} \left\| \mathbf{r}_i \times \mathbf{F}_i \right\|}
\end{eqnarray}
}
\revadd{
where $m$ is the object mass, $\mathbf{g}$ is the gravitational acceleration,
and $\mathbf{r}_i$ is the position vector from the center of mass to the $i$-th contact point.
}

\revadd{
The resistance encountered when lifting the object in a direction $\mathbf{\hat{d}}$ can be computed using $\mathbf{F}_i$ as described below. We compute $R(\mathbf{\hat{d}})$ for both the predicted force distribution and the ground-truth point forces obtained from the simulator, and define LRDA as the IoU between the corresponding low-resistance regions.
}
\revadd{
\begin{eqnarray}
R(\mathbf{d}) &=& \sum_{i} \mathbf{F}_i \cdot \hat{\mathbf{d}}, \\
\mathcal{D}_{\text{pred}} &=& \left\{ \mathbf{d} \in \mathbb{S}^2 \;\middle|\; R_{\text{pred}}(\mathbf{d}) < \epsilon \right\}, \quad
\mathcal{D}_{\text{gt}} = \left\{ \mathbf{d} \in \mathbb{S}^2 \;\middle|\; R_{\text{gt}}(\mathbf{d}) < \epsilon \right\}, \\
\mathrm{LRDA} &=& \frac{ \left| \mathcal{D}_{\text{pred}} \cap \mathcal{D}_{\text{gt}} \right| }{ \left| \mathcal{D}_{\text{pred}} \cup \mathcal{D}_{\text{gt}} \right| }
\end{eqnarray}
}
\revadd{
In the experiments, $\delta$ was set to $0.01$, and $\epsilon$ was determined so that $\mathcal{D}_{\text{pred}}$ and $\mathcal{D}_{\text{gt}}$ each cover 20\% of the total solid angle $4\pi$.
}

\end{document}